\documentclass[onecolumn]{assets/marspaper} %

\usepackage[most]{tcolorbox}

\usepackage{amsmath}
\usepackage{amssymb}
\usepackage[utf8]{inputenc} %
\usepackage[T1]{fontenc}    %
\usepackage{hyperref}       %
\usepackage{url}            %
\usepackage{booktabs}       %
\usepackage{longtable}
\usepackage{multirow}
\usepackage{amsfonts}       %
\usepackage{nicefrac}       %
\usepackage{microtype}      %
\usepackage{xcolor}         %
\usepackage{graphicx}
\newcommand{\m}[1]{{ODE}}
\newcommand{\p}[1]{}

\usepackage{wrapfig}
\usepackage{enumitem}
\usepackage{fontawesome5}

\usepackage{amsthm}   %

\usepackage{algorithm}
\usepackage{algpseudocode}
\algrenewcommand\algorithmiccomment[1]{\hfill{\color{marsink}$\triangleright$~#1}}
\usetikzlibrary{calc}
\colorlet{odebg}{marsaccent!9}
\newcommand{\odestart}[1]{\tikz[remember picture,overlay]\coordinate(#1) at (0,0);}
\newcommand{\odestop}[1]{\tikz[remember picture,overlay]{\coordinate(#1e) at (0,0);
  \fill[odebg,rounded corners=2.5pt,blend mode=multiply]
    ($(#1)+(-5pt,0.78\baselineskip)$) rectangle ($(#1e)+(1.5pt,-0.31\baselineskip)$);
  \fill[marsaccent,blend mode=multiply]
    ($(#1)+(-5pt,0.78\baselineskip)$) rectangle ($(#1|-#1e)+(-3.8pt,-0.31\baselineskip)$);}}

\theoremstyle{remark}

\newlist{checklist}{itemize}{2}
\setlist[checklist]{label=$\square$, leftmargin=*, itemsep=1pt, topsep=3pt}

\runningtitle{ODE Policy: Robot Learning via Action Dynamics, Not Actions}%
\title{ODE Policy: Robot Learning\\via Action Dynamics, Not Actions}

\author[*,\ddag]{\href{https://jiajindou.github.io/}{\textcolor{black}{Jindou Jia}}}
\author[*]{\href{https://jingliangli.com/}{\textcolor{black}{Jingliang Li}}}
\author[*]{\href{https://kuangjizuo.com/}{\textcolor{black}{Kuangji Zuo}}}
\author[*]{\href{https://morpheus-an.github.io/}{\textcolor{black}{Tuo An}}}
\author[]{\textcolor{black}{Yuxuan Hu}}
\author[]{\textcolor{black}{Mengfei Zhao}}
\author[]{\textcolor{black}{Bofan Lyu}}
\author[]{\textcolor{black}{Jiaqi Bai}}
\author[]{\textcolor{black}{Daoyuan Zhu}}
\author[]{\textcolor{black}{Hengyu Dai}}
\author[]{\textcolor{black}{Junpeng Zhang}}
\author[\dagger]{\href{https://marsyang.site/}{\textcolor{black}{Jianfei Yang}}}

\affiliation[]{MARS Lab, Nanyang Technological University}

\contribution[\ddag]{Project Lead}
\contribution[*]{Equal Contribution}
\contribution[\dagger]{Corresponding Author}

\metadata[\href{https://ntumars.github.io/ODE/}{\faGlobe~Project Page}\qquad\href{https://github.com/NTUMARS/ODE-POLICY}{\faGithub~Code}]{}%
\correspondence{\email{jindou.jia@ntu.edu.sg}, \email{jianfei.yang@ntu.edu.sg}.}

\abstract{
Most robot policies, whether trained by imitation learning (IL) or reinforcement learning (RL), map observations to
action values at discrete control steps and leave the continuity of physical motion to be learned implicitly. We
propose ODE policy, which learns the dynamics of the action rather than the action itself. A neural ordinary
differential equation (ODE) learns the vector field of the action by matching the integrated actions to the targets
over short action chunks in IL and single control steps in RL. A lightweight diversity head captures multi-modal
behavior by randomizing the direction of the field. At deployment, each action is integrated from the previous one,
so the executed trajectory is continuous by construction. We evaluate it as a visuomotor policy, as the action head
of vision-language-action (VLA) and world action models (WAMs), and as the actor of model-based RL against 14
baselines on unimanual, bimanual, dexterous, locomotion, and loco-manipulation tasks. In IL, against flow matching, it
reaches 5.9$\times$ the success, 5.4$\times$ the generalization, and 3.9$\times$ the
mode balance after one epoch, and generates actions 10.8$\times$ faster. It mostly moves more smoothly than human demonstrations. On 14 real-world tasks across single-arm, dual-arm, dexterous, and humanoid robots, it achieves the highest
success rate. In RL, it further speeds up SuperMimic for humanoid motion tracking and smooths its actions.
Efficient in learning, data, parameters, and execution, ODE policy offers a promising path to scaling robot learning.
}

\teaser{
\includegraphics[width=0.82\linewidth]{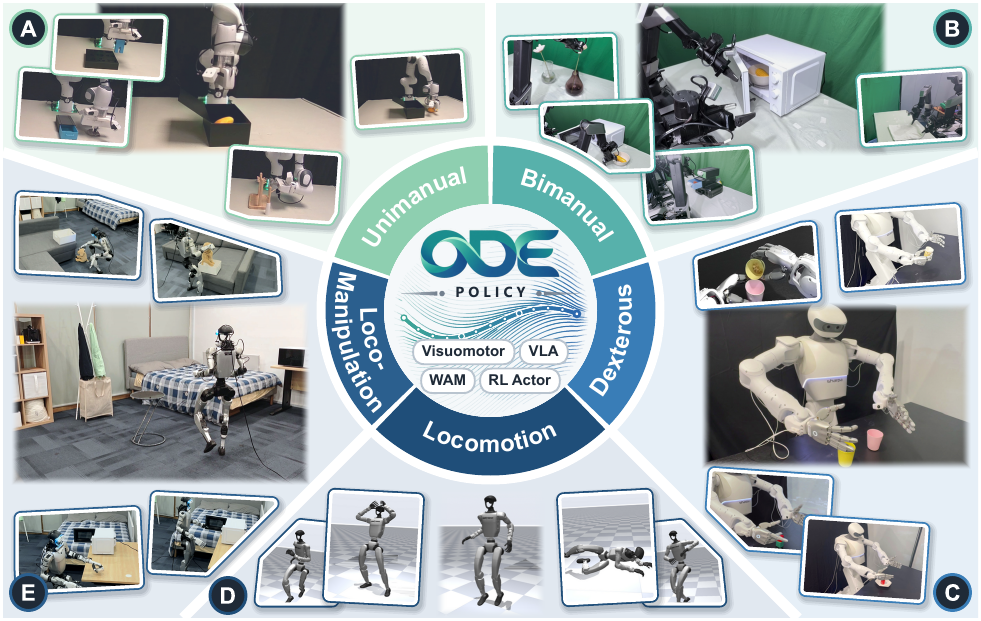}
\captionof{figure}{\textbf{\m{}~policy across embodiments and models.} The same formulation serves as a visuomotor
policy, as the action head of a VLA and a WAM, and as the
actor of RL. \textbf{(A)} Unimanual manipulation with a single arm. \textbf{(B)} Bimanual
manipulation on a dual-arm robot. \textbf{(C)} Dexterous manipulation with five-fingered hands.
\textbf{(D)} Locomotion of a simulated humanoid that tracks human motions, trained by RL. \textbf{(E)}
Loco-manipulation with a humanoid robot.}
\label{fig:teaser}
}

\begin{document}

\maketitle

\section{Introduction}
\label{sec:intro}

\begin{flushright}
{\itshape Natura non facit saltus.}\\
{\small\color{marsink}(Nature does not make jumps.)}\\[2pt]
{\small\color{marsink}--- \textsc{G.\,W. Leibniz}, 1704}
\end{flushright}

\begin{figure}[t]
\centering
\includegraphics{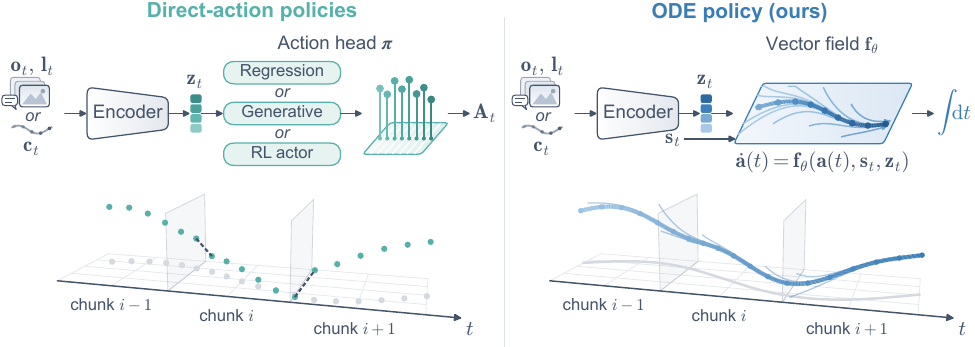}%
\caption{\textbf{Direct-action policies and \m{}~policy.} Both encode the image history
and the instruction (IL) or the reference motion (RL). \textbf{Left:} Direct-action policies, whether a regression head,
a generative head, or the actor of RL, map the embedding to action values at discrete control steps.
\textbf{Right:} \m{}~policy instead learns the dynamics of the action, a vector field $\mathbf{f}_\theta$, and integrates it in physical time, so that each action follows from
the previous one. In RL, each chunk holds a single action.}
\label{fig:concept}
\end{figure}

Robot learning has advanced rapidly in the past few years. In imitation learning (IL), visuomotor
policies~\citep{dp,zhao2023learning}, vision-language-action (VLA)
models~\citep{brohan2023rt2,kim2024openvla,black2024pi_0,intelligence2025pi05,bjorck2025gr00t}, and world action models (WAMs)~\citep{zhang2026imagewam, an2026feedback, hou2026world} now perform a wide range of manipulation tasks.
In reinforcement learning (RL), actors trained in massively parallel simulation let legged and humanoid robots walk
and track human motions~\citep{rudin2022learning,mittal2023orbit,liao2026beyondmimic}. Despite their differences,
most of these policies follow the same paradigm. They encode the history of observations and the proprioceptive
state, and map it directly to future actions at discrete time steps.

This paradigm has limits in both settings. In IL, it relies on large-scale data and compute to generalize. Yet robot
data are heterogeneous in embodiment, quality, and control interface, and cannot be collected at the scale of
internet data. Training large foundation policies is also costly. Generative policies~\citep{dp,black2024pi_0}
further need multiple network evaluations for each chunk. They react to new observations only after the whole chunk
is produced, which limits real-time execution~\citep{black2026real,bai2026flash,jia2026action}. In RL, the policy
must gather experience from a large number of interactions~\citep{rudin2022learning}. Without explicit
regularization, its actions often oscillate at high frequency. This wastes control effort, wears the actuators, and
hurts the transfer from simulation to reality~\citep{mysore2021caps,raffin2022smooth,peng2020learning}. A novel robot policy paradigm that is
\textbf{efficient in learning, data, parameters, and execution, and especially amenable to scaling,} is therefore needed.

A robot is a macroscopic classical system that moves continuously in time. The actions that command it, e.g., those
teleoperated by humans, are largely continuous as well. A demonstration is therefore largely a sampling of a
continuous action trajectory at the control frequency. Direct-action policies must learn this continuity implicitly
from data. At chunk boundaries, it is not even guaranteed, and the commands jump~\citep{black2026real}. An ordinary
differential equation (ODE)~\citep{chen2018neural} naturally describes such continuous change. Applied to the
action, an ODE can model its dynamics, a law of motion shared by every step. Integrating it yields a continuous action
trajectory by construction. Notably, animal motor cortex is thought to generate movement through such dynamics
rather than by encoding its target~\citep{churchland2012neural,shenoy2013cortical}. ODEs have long described robot
motion~\citep{ijspeert2013dmp,bahl2020ndp}. However, they fix the form of the vector field or force it to converge to an attractor. Each model is thus limited to a few motions.

In this work, we propose \textbf{\m{}~policy}. It learns the dynamics of the action rather than the action itself,
i.e., it models the policy directly as an ODE (Fig.~\ref{fig:concept}). Concretely, a neural
ODE~\citep{chen2018neural} learns the vector field of the action, conditioned on the observation. The action is then
integrated over the physical time of the robot. Integrating over a whole demonstration accumulates drift and requires
the costly \textit{adjoint} method. In IL, we therefore integrate over action chunks~\citep{dp,zhao2023learning}, a
key to current direct-action policies~\citep{pan2025much,lazzati2026}, and supervise the integrated path without any
derivative of the actions. A contractive perturbation shifts the start of each training chunk and teaches the field
to return, which improves robustness. In RL, we integrate over single control steps. At inference, each action is
integrated from the previous one, across chunk boundaries in IL and across control steps in RL.
Direct-action policies, in contrast, often jump or pause at chunk boundaries in IL and oscillate between steps in RL. Finally, a lightweight diversity head adds a sampled offset to the vector
field to retain multi-modal behavior~\citep{dp,jia2026mars}.

We validate \m{}~policy across the mainstream IL models and in RL (Fig.~\ref{fig:teaser}). We instantiate it as a
visuomotor policy, as the action expert of the $\pi_{0.5}$ VLA~\citep{intelligence2025pi05}, as the action head
of the ImageWAM~\citep{zhang2026imagewam} WAM, and as the actor of SuperMimic~\citep{liao2026supermimic} for
whole-body motion tracking of a humanoid. We further deploy it on a real single-arm robot, a dual-arm robot, a robot
with dexterous hands, and a humanoid, where it replaces the action head of GR00T N1.7~\citep{bjorck2025gr00t}. The
real-robot tasks include high-precision, long-horizon, deformable, multi-modal, dexterous, and loco-manipulation tasks. We compare with
14 state-of-the-art baselines, including generative and regression-based IL methods. Within each setting, all
methods share the same backbone, data, and training schedule.

Experiments show that \m{}~policy \textbf{learns remarkably fast}. After one epoch, it succeeds in 79.2\% of the
LIBERO episodes, against at most 43.2\% for the baselines. As the action head of $\pi_{0.5}$, it succeeds in 96.0\%
against 75.5\% for its flow matching expert. As that of ImageWAM, it succeeds in 82.6\% against 24.6\% for the
original head. It matches flow matching with less than half of the demonstrations. As the actor of SuperMimic, it reaches a higher return early in training. On 14 tasks across single-arm, dual-arm, dexterous, and humanoid
robots, it achieves the highest success rate on every task. It also \textbf{generates actions fast and smoothly}.
A chunk takes 10.8 times less time than with flow matching in simulation, and 3.5 to 27 times less than with the
diffusion and flow matching baselines on real robots. On real robots, its motion is smoother than the human demonstrations in most cases. Moreover, it \textbf{generalizes robustly} to unseen perturbations in simulation, and it transfers
dexterous skills learned from public data of another lab to our real robot. It also \textbf{keeps multi-modal
behavior}, both on a navigation task and on a real bimanual task, where it uses both arm assignments. Finally, the
learned field supports data without action labels and arbitrary control rates.

\section{Related Work}
\label{sec:related}

\subsection{ODE in Robot Policies}
\label{sec:related_ds}

\textbf{{Action dynamics as ODEs.}} Representing robot motion as a dynamical system has a long history, from
dynamical movement primitives~\citep{ijspeert2013dmp} and stable dynamical
systems~\citep{khansari2011seds} to neural vector fields with contraction or stability
guarantees~\citep{beikmohammadi2024ncds,sochopoulos2024stable}. Neural dynamic
policies~\citep{bahl2020ndp,totsila2023andp} further predict the parameters of such systems from
images. However, the fixed form or the stability constraint limits each system to a few motions
toward an attractor, mostly on single tasks. \m{}~policy instead places no structural constraint on
the vector field and learns one field for large-scale tasks.

\textbf{{State dynamics as ODEs.}} A conventional paradigm in robot learning is to learn the
state dynamics of the robot or its environment and then design a policy or analytical controller upon it. Learned
dynamics, often in the form of neural ODEs, are used for model-based reinforcement learning~\citep{yildiz2021continuous}
and for feedback controllers that compensate for model
uncertainty~\citep{jia2024feedback,jia2025FORESEERR,jia2026physics}. Continuous-time
networks~\citep{hasani2022cfc} likewise evolve the hidden state of a policy by an ODE. In these
works, the ODE describes the state of the system or of the network, and the action is still
produced by a separate controller or output layer. \m{}~policy instead learns the ODE of the action
itself.

\textbf{{Generative processes as ODEs.}} Diffusion and flow matching policies~\citep{dp,black2024pi_0}
integrate over a generation time without physical meaning. Streaming flow
policy~\citep{jiang2025streaming} aligns the flow time with the execution time, but it is still
trained by regressing velocities derived from the demonstrations and regenerates each chunk over a
rescaled time interval. Chronos policy~\citep{wang2026chronos} encodes the physical times of observations and
actions in a diffusion transformer, but time remains an input, and actions are still generated at a fixed rate.
In contrast, \m{}~policy integrates over physical time itself, is trained through the integrated path without
velocity labels, and continues each chunk from the previous one.

\subsection{Action Chunking}
\label{sec:related_chunk}

Predicting a chunk of actions per observation was popularized by action chunking transformer (ACT)~\citep{zhao2023learning} and
diffusion policy~\citep{dp}, and is now standard in large foundation
models~\citep{black2024pi_0,intelligence2025pi05,bjorck2025gr00t}. Recent analyses show that
chunking is a key factor in the performance of current IL policies~\citep{lazzati2026,pan2025much}.
Executing a chunk open-loop often outperforms updating the action at every observation~\citep{park2026bcmystery},
which contrasts with the classical control experience that a higher closed-loop rate is better.
However, independently predicted chunks may be inconsistent at their boundaries, which is mitigated
by temporal ensembling~\citep{zhao2023learning} or by conditioning each chunk on the committed
actions of the previous one~\citep{black2026real}.

\m{}~policy also builds on chunking for IL, but uses the chunk as the unit of integration. During training, it
integrates the vector field over one chunk, so the gradient passes through a few solver steps without the
\textit{adjoint} method. At inference, it integrates each action from the previous one, which keeps the actions
continuous across chunks and executes each action without waiting for a whole chunk. Some policies predict actions
relative to the current pose~\citep{chi2024universal}, but still output each action of a chunk as a separate value.
In \m{}~policy, all actions of a chunk follow from one vector field evaluated along the path.

\subsection{Efficient Policies}
\label{sec:related_efficient}

\textbf{{Efficient IL policies.}} Generative policies require multiple network evaluations per
chunk. Consistency models~\citep{Consistency,zhang2025flowpolicy}, self-distillation~\citep{li2026ofp},
and mean flows~\citep{geng2026mean,fang2025omp} reduce the sampling to one or a few steps, and
informed source distributions shorten the transport
path~\citep{warmstarts,dai2026where}. VITA~\citep{gao2025vita}, action-to-action flow matching (A2A)~\citep{jia2026action,zuo2026deltaflow}, and
FLASH~\citep{bai2026flash} further replace the flow source with image observations or historical
actions, which improves both learning efficiency and inference speed. These methods improve the efficiency of IL
within the conventional direct-action paradigm. \m{}~policy instead changes the paradigm to an ODE, which
exploits the inherent continuity of physical motion to improve efficiency.

\textbf{{Efficient RL policies.}} In RL, sample efficiency is mostly improved on the side of the
critic and the update, e.g., by high update-to-data ratios with ensembled or regularized
critics~\citep{chen2021redq,hiraoka2022droq,bhatt2024crossq}, scaled
networks~\citep{lee2025simba,kim2026flashsac}, or learned world models~\citep{hansen2024tdmpc2}.
SuperMimic~\citep{liao2026supermimic} learns a differentiable world model and trains the actor by backpropagating
analytic gradients through short imagined rollouts. It surpasses a state-of-the-art motion tracker after 3 hours
on one GPU with more than 1000 times fewer GPU hours. These methods keep the actor as a direct map from states to
actions. \m{}~policy instead changes the actor itself, and further speeds up SuperMimic as a drop-in replacement of
its actor.

\section{Preliminaries}
\label{sec:preliminaries}

This section first formalizes the direct-action policies that \m{}~policy replaces, in both IL and RL, and then
introduces the neural ODE on which \m{}~policy is built.

\subsection{Direct-Action Policy}
\label{sec:prelim_il}

Let $\mathbf{o}_t\in\mathcal{O}$ denote the visual observation in the image space $\mathcal{O}$,
$\mathbf{s}_t\in\mathbb{R}^{d_s}$ the proprioceptive state, and $\mathbf{a}_t\in\mathbb{R}^{d_a}$ the action sent to the low-level controller at control step $t\in\mathbb{N}$, where
$d_s,d_a\in\mathbb{N}$ are the state and action dimensions. We
write $\mathbf{c}_t=(\mathbf{o}_t,\,\mathbf{l}_t)$ for the condition, with a language
instruction $\mathbf{l}_t\in\mathcal{T}$ in the space of instruction texts $\mathcal{T}$. The observation and the state
usually include a short history.
A conventional policy maps the condition and the state to the action value at each control step,
\begin{equation}
    \mathbf{a}_t\sim\boldsymbol{\pi}(\cdot\mid\mathbf{c}_t,\mathbf{s}_t),
    \label{eq:discrete}
\end{equation}
where each action is predicted as a separate value. The model imposes no relation between consecutive
actions. Any such relation, such as the continuity of the motion, must be learned implicitly from data.
We call such policies \textit{direct-action policies}, since the network outputs the action values themselves.

\textbf{{IL policies.}} IL fits $\boldsymbol{\pi}$ to a dataset $\mathcal{D}$ of $N\in\mathbb{N}$ demonstrated trajectories
$\{(\mathbf{o}_t,\mathbf{s}_t,\mathbf{l}_t,\mathbf{a}_t)\}$. Most IL policies follow action
chunking~\citep{zhao2023learning,dp} and predict a chunk of $H\in\mathbb{N}$ future actions at once,
\begin{equation}
    \mathbf{A}_t=[\mathbf{a}_{t},\dots,\mathbf{a}_{t+H-1}]\sim\boldsymbol{\pi}(\cdot\mid\mathbf{c}_t,\mathbf{s}_t),
    \label{eq:discrete_chunk}
\end{equation}
with $\mathbf{A}_t\in\mathbb{R}^{H\times d_a}$, which is executed before the next observation is processed. Each action
$\mathbf{a}_{t+k}$ with $k\in\{0,\dots,H-1\}$ is still a separate value, within and across chunks, and $H=1$
recovers Eq.~\eqref{eq:discrete}. Explicit regression maps
$(\mathbf{c}_t,\mathbf{s}_t)$ directly to $\mathbf{A}_t$ under an $\ell_1$ or $\ell_2$
loss~\citep{pomerleau1989alvinn,zhao2023learning}, which is simple but averages multi-modal
demonstrations. Generative policies instead sample $\mathbf{A}_t$ from a learned distribution.
Diffusion policies~\citep{dp,ddpm} iteratively denoise a chunk drawn from \textit{Gaussian} noise, and flow
matching policies~\citep{flowmatching,liu2023rectified,black2024pi_0,jia2026action} integrate a
deterministic vector field that transports a source sample to $\mathbf{A}_t$ over a time
$\tau\in[0,1]$.

Notably, although
diffusion and flow matching solve a differential equation at inference, $\tau$ is a
\emph{generation time} without physical meaning. The intermediate sample is not a state the robot
can occupy, and each chunk is generated anew, independently of where the previous chunk ended.

\textbf{{RL policies.}} RL instead trains $\boldsymbol{\pi}$ through interaction to maximize the expected
discounted return. Without images, $\mathbf{s}_t$ holds the state history, and for motion tracking the condition
$\mathbf{c}_t$ holds the reference motion. The actor typically outputs a single action at each control step as in
Eq.~\eqref{eq:discrete}. Model-based methods such as SuperMimic~\citep{liao2026supermimic} update the actor by
backpropagating the return of rollouts imagined in a differentiable world model.

\subsection{Neural ODE}
\label{sec:prelim_node}

A neural ODE~\citep{chen2018neural,jia2024feedback} parameterizes the time derivative
of a continuous state $\mathbf{x}(t)\in\mathbb{R}^{n}$, $n\in\mathbb{N}$, at time $t\in\mathbb{R}$ with a neural network
$\mathbf{g}_\phi:\mathbb{R}^{n}\times\mathbb{R}\to\mathbb{R}^{n}$ with parameters $\phi$, and obtains its trajectory
from $t_0$ to $t_1$ ($t_0,t_1\in\mathbb{R}$, $t_0<t_1$) by integration,
\begin{equation}
    \dot{\mathbf{x}}(t)=\mathbf{g}_\phi\big(\mathbf{x}(t),t\big),\qquad
    \mathbf{x}(t_1)=\mathbf{x}(t_0)+\int_{t_0}^{t_1}\mathbf{g}_\phi\big(\mathbf{x}(t),t\big)\,\mathrm{d}t ,
    \label{eq:node}
\end{equation}
where the integral is evaluated by a numerical solver (e.g., the \textit{Euler} method or \textit{Runge--Kutta} methods such as RK2 and RK4) and the network is trained by
back-propagating a loss on the integrated states. The gradient is obtained either directly
through the solver steps or by the \textit{adjoint} method~\citep{chen2018neural}, which integrates an
augmented ODE backward in time with constant memory at the cost of additional integration. The
network only specifies
the local rate of change, while the continuity and smoothness of the trajectory come from
integration. Earlier robot policies built on dynamical systems, such as movement
primitives~\citep{ijspeert2013dmp}, stable dynamical systems~\citep{khansari2011seds}, and neural
dynamic policies~\citep{bahl2020ndp}, fix the form of the vector field in advance and learn only its
parameters. In this work, we instead let the network itself be the vector field of the action,
integrated over the physical time of the robot.

\section{Method}
\label{sec:method}

In this section, we propose \m{}~policy, which models how the action evolves over the physical time of the robot as
an ODE (Fig.~\ref{fig:concept}). We present it mainly for IL. We first formulate the policy as a vector field
of the action (Section~\ref{sec:method_field}), then describe how it is trained from demonstrations
(Section~\ref{sec:method_loss}) and executed at inference (Section~\ref{sec:method_infer}).
Section~\ref{sec:method_side} adds an optional diversity head for multi-modal behavior. Finally,
Section~\ref{sec:method_rl} describes how the same field serves as the actor in RL.

\subsection{Vector Field Formulation}
\label{sec:method_field}

\textbf{{Vector field.}} We encode the image history and the instruction into an embedding $\mathbf{z}_t\in\mathbb{R}^{d_z}$, $d_z\in\mathbb{N}$, and parameterize the velocity of the action with a neural network~\citep{chen2018neural},
\begin{equation}
    \dot{\mathbf{a}}(t)=\mathbf{f}_\theta\big(\mathbf{a}(t),\mathbf{s}_t,\mathbf{z}_t\big),
    \label{eq:ode_policy}
\end{equation}
where $\mathbf{f}_\theta:\mathbb{R}^{d_a}\times\mathbb{R}^{d_s}\times\mathbb{R}^{d_z}\to\mathbb{R}^{d_a}$, $t\in\mathbb{R}_{\ge0}$ is the physical time of the
robot in units of control steps, $\mathbf{a}(t)\in\mathbb{R}^{d_a}$ is an action the robot is commanded to reach
and $\dot{\mathbf{a}}$ is its velocity. The proprioception $\mathbf{s}_t$ enters the field directly rather than being encoded into $\mathbf{z}_t$ together with $\mathbf{c}_t$ as in conventional policies (ablated in Section~\ref{sec:exp_ablation}).

\textbf{{Chunked integration.}} At a decision step $t$, the policy computes $\mathbf{z}_t$ once and
holds $\mathbf{z}_t$ and $\mathbf{s}_t$ fixed over the chunk. Starting from the last action
$\hat{\mathbf{a}}_{t-1}\in\mathbb{R}^{d_a}$, it integrates Eq.~\eqref{eq:ode_policy} with a numerical solver,
\begin{equation}
    \hat{\mathbf{a}}_{t+k}=\hat{\mathbf{a}}_{t+k-1}
    +\int_{t+k-1}^{t+k}\mathbf{f}_\theta\big(\mathbf{a}(t'),\mathbf{s}_t,\mathbf{z}_t\big)\,\mathrm{d}t',
    \qquad k=0,\dots,H-1,
    \label{eq:chunk}
\end{equation}
where $t'\in\mathbb{R}$ is the integration variable, and we obtain the chunk
$\hat{\mathbf{A}}_t=[\hat{\mathbf{a}}_{t},\dots,\hat{\mathbf{a}}_{t+H-1}]\in\mathbb{R}^{H\times d_a}$ at the same
steps as $\mathbf{A}_t$ in Eq.~\eqref{eq:discrete_chunk}.
In contrast to Eqs.~\eqref{eq:discrete} and~\eqref{eq:discrete_chunk}, each action is obtained from the previous one by integrating its
rate of change, so consecutive actions are related by construction.
Notably, the field takes the current action $\mathbf{a}(t)$ as input, so the change at each step depends on
where the action has arrived. A conventional policy instead outputs the actions or increments of a chunk as $H$ separate values,
whose relation must be learned from data.
Holding the condition fixed matches action chunking~\citep{zhao2023learning,dp}, where the
observation is processed once per chunk~\citep{lazzati2026}.

Note that the integrated state is the desired action, i.e., the target sent to the low-level
controller, rather than the measured state of the robot. Since the low-level controller, e.g., a
proportional-derivative (PD) or operational space controller~\citep{khatib1987unified}, tracks its target imperfectly, the robot state always lags
behind the target~\citep{zhao2026apex}. For a delta controller~\citep{zhu2020robosuite}, the desired action is the current pose plus the scaled command,
from which the command is recovered exactly at execution. When only the measured state is available,
e.g., data collected by UMI~\citep{chi2024universal} or from human videos~\citep{kareer2025emergence}, we discuss how the velocity of the field naturally compensates for the lag in
Section~\ref{sec:discussion}.

\subsection{Training}
\label{sec:method_loss}

\textbf{{Path loss.}} Given a demonstration segment, we start from the demonstrated action
$\mathbf{a}_{t-1}$, integrate Eq.~\eqref{eq:chunk} for $H$ steps, and match the integrated path,
\begin{equation}
    \mathcal{L}_{\text{path}}=\frac{1}{H}\sum_{k=0}^{H-1}
    \big\|\hat{\mathbf{a}}_{t+k}-\mathbf{a}_{t+k}\big\|_2^2 .
    \label{eq:path_loss}
\end{equation}
Since the chunk is usually short, the gradient is back-propagated directly through the solver steps without
the \textit{adjoint} method~\citep{chen2018neural}. The field is supervised only through the integrated path and never regressed
against velocity labels as in \citet{jiang2025streaming}, which may amplify the jitter of noisy demonstrations.

\textbf{{Contractive perturbation.}} The path loss supervises the vector field only along the
demonstrated trajectories. At execution, however, the policy integrates from its own previous
output, so any error carries over to the next step and the next chunk, and the state gradually
leaves the demonstrations. To improve robustness, we
perturb the start point of a segment during training and require the integrated path to return to the
demonstration,
\begin{equation}
    \mathcal{L}_{\text{path}}^{\boldsymbol{\delta}}=\frac{1}{H}\sum_{k=0}^{H-1}
    \big\|\hat{\mathbf{a}}_{t+k}(\mathbf{a}_{t-1}+\boldsymbol{\delta})
    -(\mathbf{a}_{t+k}+\rho^{k+1}\boldsymbol{\delta})\big\|_2^2,
    \qquad \boldsymbol{\delta}\sim\mathcal{N}\big(\mathbf{0},\mathrm{diag}(\boldsymbol{\sigma}_\delta^2)\big),
    \label{eq:perturb_loss}
\end{equation}
where $\hat{\mathbf{a}}_{t+k}(\cdot)$ denotes the path of Eq.~\eqref{eq:chunk} integrated from the
given start point, $\boldsymbol{\delta}\in\mathbb{R}^{d_a}$ is the perturbation with scale
$\boldsymbol{\sigma}_\delta\in\mathbb{R}^{d_a}_{>0}$, and $\rho\in(0,1)$ is a decay rate. Each segment
is perturbed with probability $p\in[0,1]$, and $\boldsymbol{\delta}=\mathbf{0}$ otherwise, which
recovers Eq.~\eqref{eq:path_loss}. Since the target deviation $\rho^{k+1}\boldsymbol{\delta}$ shrinks
geometrically, the field learns to pull nearby states back to the demonstration at a rate set by
$\rho$, i.e., a local contraction~\citep{beikmohammadi2024ncds} around the demonstrations. We therefore call it
\textit{contractive perturbation}. It relies on the field taking the action as input. Standard direct-action policies do not take
the commanded action as input, so such a perturbation has no counterpart for them. The perturbation requires neither additional data nor additional computation at inference (ablated in Fig.~\ref{fig:ablation2}(A)).

The losses above are defined on a single demonstration segment
$\xi_t=(\mathbf{c}_t,\mathbf{s}_t,\mathbf{a}_{t-1:t+H-1})$. The total objective averages them over the segments in $\mathcal{D}$
and the perturbations,
\begin{equation}
    \mathcal{L}(\theta,\psi)=\mathbb{E}_{\xi_t\sim\mathcal{D},\,\boldsymbol{\delta}}
    \big[\mathcal{L}_{\text{path}}^{\boldsymbol{\delta}}(\xi_t)+\underbrace{\mathcal{L}_{\text{div}}(\xi_t)}_{\text{\textit{optional}}}\big],
    \label{eq:total_loss}
\end{equation}
which is minimized over mini-batches of segments, where the \textit{optional} diversity loss
$\mathcal{L}_{\text{div}}$ is introduced in Section~\ref{sec:method_side}. Algorithm~\ref{alg:ode_train}
contrasts this training with that of a direct-action policy.

\begin{algorithm}[!tbp]
\caption{\textbf{IL training pseudocode.}}
\label{alg:ode_train}
\small
\begin{minipage}[t]{0.45\linewidth}
\textbf{Direct-action policy}
\begin{algorithmic}[1]
\For{each segment $(\mathbf{c}_t,\mathbf{s}_t,\mathbf{A}_t)$ in $\mathcal{D}$}
  \State update $\theta$ to fit $\mathbf{A}_t$ \Comment{regressed or generated}
\EndFor
\end{algorithmic}
\end{minipage}\hfill
\begin{minipage}[t]{0.51\linewidth}
\textbf{\m{}~policy}
\begin{algorithmic}[1]
\For{each segment $(\mathbf{c}_t,\mathbf{s}_t,\mathbf{a}_{t-1:t+H-1})$ in $\mathcal{D}$}
  \State \odestart{odetr}$\hat{\mathbf{a}}_{t-1}\gets\mathbf{a}_{t-1}+\boldsymbol{\delta}$ \Comment{contractive perturbation}
  \State integrate $\mathbf{f}_\theta$ for $H$ steps from $\hat{\mathbf{a}}_{t-1}$
  \State update $\theta$ with $\mathcal{L}_{\text{path}}^{\boldsymbol{\delta}}$ through the solver\hfill\odestop{odetr}
\EndFor
\end{algorithmic}
\end{minipage}
\end{algorithm}

\subsection{Inference}
\label{sec:method_infer}

At inference, each action within a chunk is obtained from the previous one by a single integration step
(Algorithm~\ref{alg:ode}). Across
chunks, we use \textit{seamless chunking}, i.e., each chunk is integrated with Eq.~\eqref{eq:chunk} from the last
action of the previous one, $\hat{\mathbf{a}}_{t-1}$, rather than from the measured state (ablated in
Fig.~\ref{fig:ablation2}(B, C)). This inference gives three benefits over direct-action policies.
\textbf{{(i) Smooth execution.}} Unlike independently generated chunks~\citep{zhao2023learning,black2026real},
successive chunks join without jumps, so the executed trajectory is a single integral curve whose condition is
refreshed every $H$ steps. \textbf{{(ii) Streaming execution.}} Each action is sent to the controller as soon as it
is integrated, whereas direct-action policies must generate the whole chunk before executing its first action.
\textbf{{(iii) Free integration step.}} Since the field is defined in continuous time, the integration step can be
chosen at deployment, e.g., to match a higher control rate, whereas direct-action policies output actions at the fixed
step of their training data (Fig.~\ref{fig:discussion}(C)).

\begin{algorithm}[!tbp]
\caption{\textbf{IL inference pseudocode.}}
\label{alg:ode}
\small
\begin{minipage}[t]{0.45\linewidth}
\textbf{Direct-action policy}
\begin{algorithmic}[1]
\For{each decision step $t=0,H,2H,\dots$}
  \State $\hat{\mathbf{A}}_t\sim\boldsymbol{\pi}(\cdot\mid\mathbf{c}_t,\mathbf{s}_t)$ \Comment{a new chunk}
  \State execute $\hat{\mathbf{A}}_t$ \Comment{once it is complete}
\EndFor
\end{algorithmic}
\end{minipage}\hfill
\begin{minipage}[t]{0.51\linewidth}
\textbf{\m{}~policy}
\begin{algorithmic}[1]
\For{each decision step $t=0,H,2H,\dots$}
  \For{$k=0,\dots,H-1$}
    \State \odestart{odein}$\hat{\mathbf{a}}\gets\textsc{Step}(\mathbf{f}_\theta,\hat{\mathbf{a}},\mathbf{s}_t,\mathbf{z}_t)$, send $\hat{\mathbf{a}}$ \Comment{streaming\odestop{odein}}
  \EndFor
\EndFor \hfill\odestart{odenx}{\color{marsink}$\triangleright$~next chunk starts from $\hat{\mathbf{a}}$}\odestop{odenx}
\end{algorithmic}
\vspace{2pt}\end{minipage}
\end{algorithm}

\subsection{Diversity Head for Multi-Modality}
\label{sec:method_side}

\begin{figure}[!tbp]
\centering
\includegraphics{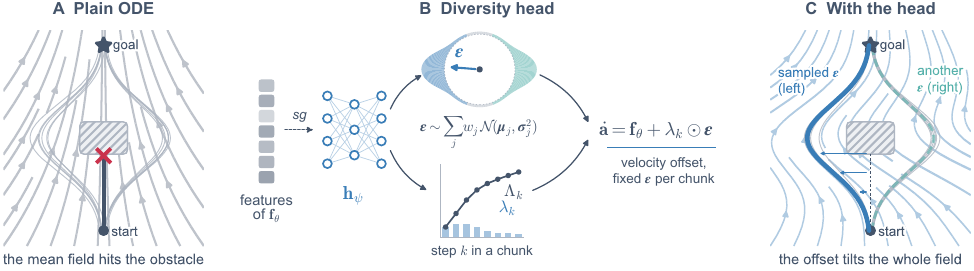}%
\caption{\textbf{Diversity head for multi-modality.} Demonstrations (gray) pass an obstacle on the left or the
right. \textbf{(A)} The plain field averages them and hits the obstacle. \textbf{(B)} The head samples a direction
$\boldsymbol{\epsilon}$ with one mode per route and a rate $\boldsymbol{\lambda}_k$ per step, and adds
$\boldsymbol{\lambda}_k\odot\boldsymbol{\epsilon}$ to the velocity. \textbf{(C)} The offset tilts the whole field, so
each sample takes one route.}
\label{fig:diversity}
\end{figure}

Multi-modal behavior means that several distinct actions are valid in the same situation, e.g., passing an
obstacle on the left or on the right~\citep{dp,jia2026mars,ma2026moss}. A deterministic field $\mathbf{f}_\theta$ fits the
mean velocity of the demonstrations and thus averages such modes, e.g., heading straight into the obstacle
(Fig.~\ref{fig:diversity}(A)).

Since the policy is a vector field, a mode is decided by the direction in which the integration sets off, and
randomizing this direction is a natural way to select a mode. We therefore keep the ODE and add a lightweight diversity head
$\mathbf{h}_\psi$ with parameters $\psi$, which steers each chunk toward one mode by a sampled velocity offset (Fig.~\ref{fig:diversity}(B, C)). At the start of a chunk, $\mathbf{h}_\psi$ reads
the hidden features of $\mathbf{f}_\theta$ with the gradient stopped and outputs a direction and a scale. The
direction is an offset $\boldsymbol{\epsilon}\in\mathbb{R}^{d_a}$ drawn from a $K$-component ($K\in\mathbb{N}$)
\textit{Gaussian} mixture~\citep{bishop1994mdn}, whose components $j\in\{1,\dots,K\}$ cover the modes with
weights $w_j\in[0,1]$ summing to one, means $\boldsymbol{\mu}_j\in\mathbb{R}^{d_a}$, and standard deviations
$\boldsymbol{\sigma}_j\in\mathbb{R}^{d_a}_{>0}$. The scale is a non-negative rate
$\boldsymbol{\lambda}_k\in\mathbb{R}^{d_a}_{\ge0}$ for each step $k=0,\dots,H-1$. The chunk is then integrated with
the added velocity,
\begin{equation}
    \dot{\mathbf{a}}(t')=\mathbf{f}_\theta\big(\mathbf{a}(t'),\mathbf{s}_t,\mathbf{z}_t\big)
    +\boldsymbol{\lambda}_k\odot\boldsymbol{\epsilon},
    \qquad t'\in[t+k-1,\,t+k],
    \label{eq:side_exec}
\end{equation}
where $\odot$ denotes the element-wise product. Since $\boldsymbol{\epsilon}$ is fixed over the chunk and each
step lasts one control step, integrating the added velocity shifts the path at step $k$ by about
$\boldsymbol{\Lambda}_k\odot\boldsymbol{\epsilon}$ with $\boldsymbol{\Lambda}_k=\sum_{i=0}^{k}\boldsymbol{\lambda}_i\in\mathbb{R}^{d_a}_{\ge0}$,
and $\boldsymbol{\epsilon}=\mathbf{0}$ recovers the plain ODE. The head should thus produce shifts that match
how the demonstrations deviate from the plain path.

\textbf{{Diversity loss.}} To this end, let
$\mathbf{e}_k=\mathrm{sg}(\mathbf{a}_{t+k}-\hat{\mathbf{a}}_{t+k})\in\mathbb{R}^{d_a}$ denote the deviation of the
demonstration from the plain path at step $k$, where
$\mathrm{sg}$ stops the gradient. The scale $\boldsymbol{\Lambda}_k$ is fitted by the \textit{Gaussian} negative
log-likelihood of $\mathbf{e}_k$, and the mixture by the likelihood of the offset that best explains the
deviation, $\boldsymbol{\epsilon}^{*}=\arg\min_{\boldsymbol{\epsilon}}\sum_{k}\|\mathbf{e}_k-\boldsymbol{\Lambda}_k\odot\boldsymbol{\epsilon}\|_2^2$.
Since the problem decouples per dimension, it has the closed form
$\boldsymbol{\epsilon}^{*}=\big(\sum_{k}\boldsymbol{\Lambda}_k\odot\mathbf{e}_k\big)\oslash\big(\sum_{k}\boldsymbol{\Lambda}_k\odot\boldsymbol{\Lambda}_k\big)$,
where $\oslash$ denotes the element-wise division. The diversity loss is then
\begin{equation}
    \mathcal{L}_{\text{div}}=\frac{1}{H}\sum_{k=0}^{H-1}
    \Big(\tfrac{1}{2}\big\|\mathbf{e}_k\oslash\boldsymbol{\Lambda}_k\big\|_2^2
    +\mathbf{1}^{\top}\log\boldsymbol{\Lambda}_k\Big)
    -\log\sum_{j=1}^{K}w_j\,
    \mathcal{N}\big(\boldsymbol{\epsilon}^{*};\boldsymbol{\mu}_j,\mathrm{diag}(\boldsymbol{\sigma}_j^2)\big),
    \label{eq:side_loss}
\end{equation}
where $\log$ is applied element-wise,
$\mathbf{1}\in\mathbb{R}^{d_a}$ is the all-ones vector, and $\mathcal{N}(\cdot;\boldsymbol{\mu},\boldsymbol{\Sigma})$ is the
\textit{Gaussian} density. Since both its input and its target are detached, the head leaves the gradient of
$\mathbf{f}_\theta$ unchanged.

\textbf{{When to use it.}} When the same observation is followed by distinct actions in the demonstrations, the
diversity head preserves these modes (Section~\ref{sec:exp_multimodal}). However, such strong multi-modality, with
clearly distinct actions, occurs in only a small part of robot tasks. On LIBERO, for example, the actions for the same observation differ only slightly. Without
the head, \m{}~policy still learns efficiently and reaches 85.3\% after one epoch (Table~\ref{tab:e12_div}),
possibly because regression can already produce slight multi-modal behavior with a sufficient \textit{Lipschitz}
constant~\citep{pan2025much}. For completeness and a fair comparison, all our IL experiments nevertheless use the
head, sampling $\boldsymbol{\epsilon}$ within the drawn component on the multi-modal tasks and using the component
mean $\boldsymbol{\mu}_j$ elsewhere (ablated in Appendix~\ref{sec:app_ablation}).

\subsection{Reinforcement Learning}
\label{sec:method_rl}

The same field also serves as the actor in RL. Without demonstrations, the losses above are not used. The actor
acts at every control step ($H=1$) and takes one \textit{Euler} step from the previous action,
\begin{equation}
    \hat{\mathbf{a}}_{t}=\hat{\mathbf{a}}_{t-1}+\mathbf{f}_\theta\big(\hat{\mathbf{a}}_{t-1},\mathbf{s}_t,\mathbf{z}_t\big),
    \label{eq:rl_step}
\end{equation}
i.e., Eq.~\eqref{eq:chunk} with $H=1$ and one \textit{Euler} step, where $\mathbf{z}_t$ embeds the reference motion
$\mathbf{c}_t$ instead of the images. The field is trained through this step by the objective of the RL algorithm, while the critic and the
algorithm stay unchanged. This step changes how the gradient flows in SuperMimic, which backpropagates the return
through 32 imagined steps. Since $\partial\hat{\mathbf{a}}_t/\partial\hat{\mathbf{a}}_{t-1}$ contains the identity,
the gradient reaches early actions along the chain of actions without decay (measured in
Section~\ref{sec:exp_rl}).

\section{Experimental Setup}
\label{sec:exp}
\label{sec:exp_setup}

We evaluate \m{}~policy on four questions. \textbf{(1)} Does it learn faster and from less data than
conventional direct-action policies? \textbf{(2)} Does it generate actions faster and more
smoothly? \textbf{(3)} Does it generalize robustly to unseen perturbations? \textbf{(4)} Can it represent
multi-modal behavior? Section~\ref{sec:exp_sim} answers them in simulation, Section~\ref{sec:exp_real} verifies
these advantages on four real robots, and Section~\ref{sec:exp_ablation} ablates each component of \m{}~policy.
This section describes the setup shared by all three.

\textbf{{Benchmarks.}} For visuomotor IL, we train one model on the four LIBERO suites~\citep{liu2023libero}
and test its generalization zero-shot on LIBERO-plus~\citep{fei2025liberoplus}. As the action head of a
VLA and a WAM, we fine-tune
$\pi_{0.5}$~\citep{intelligence2025pi05} and ImageWAM~\citep{zhang2026imagewam} on the same suites and test them on
LIBERO and on a shared set of LIBERO-plus instances. Multi-modality is tested on a
2D navigation task with four routes. On a Franka Research 3 arm, a dual-arm Galaxea R1 Lite, a Sharpa North robot with Sharpa
dexterous hands, and a Unitree G1 humanoid with Inspire DFX dexterous hands, we test 14 tasks that are
high-precision, long-horizon, deformable, multi-modal, dexterous, or loco-manipulation, three of them trained on
data from another lab. For RL, we replace the actor of SuperMimic~\citep{liao2026supermimic}, which tracks the 40
LAFAN1 motion clips~\citep{harvey2020robust} with a simulated Unitree G1 humanoid.
Details of all benchmarks are
given in Appendix~\ref{sec:app_setup}.

\textbf{{Baselines.}} For IL, we compare with six visuomotor policies that differ only in the action head,
namely flow matching~\citep{flowmatching} with a DiT~\citep{Peebles_2023_ICCV} or a 1D UNet head (FM-DiT and
FM-UNet), diffusion~\citep{ddpm,dp} with the same two heads (DDPM-DiT and DDPM-UNet), A2A with a DiT head (A2A-DiT)~\citep{jia2026action}, and ACT~\citep{zhao2023learning}. All of them share the image and
language encoder and have about 24M parameters outside it. On the real robots, the baselines are drawn from FM-DiT,
DDPM-UNet, and the same UNet sampled in 16 DDIM steps~\citep{song2021ddim} (DDIM-UNet), with A2A-DiT added on the
single-arm robot. On the humanoid, the baseline is the original action head of GR00T N1.7~\citep{bjorck2025gr00t}. On the navigation task, we add MARS~\citep{jia2026mars}, a policy with adaptive multi-modal ability.
In $\pi_{0.5}$, we compare with its original flow matching action expert, initialized either from the pretrained
weights of $\pi_{0.5}$ (FM) or randomly (FM-R). In ImageWAM, we compare with its
original flow matching head and an A2A head~\citep{zuo2026deltaflow}.

In RL, the baseline is the original
direct-action actor of SuperMimic, and \m{}~policy keeps its network and only reads its output as the velocity of
the action. We also compare with PPO~\citep{schulman2017ppo}, which learns without a world model.

\textbf{{Implementation.}} \m{}~policy uses a FiLM-MLP vector field~\citep{perez2018film} (architectures compared in Appendix~\ref{sec:app_ablation}),
chunks of $H=8$ steps integrated with RK4, the contractive perturbation, and the diversity head with
$K=12$ components, whose component means are used on LIBERO. The real-robot tasks use the same $H=8$ and RK4, except on the humanoid.
On LIBERO, the binary gripper command is predicted by a separate classification head, a two-layer MLP of 0.27M parameters on the condition, and integrating it with the
field instead is ablated in Section~\ref{sec:disc_signals}. In $\pi_{0.5}$, it replaces the action expert of 430M
parameters by a learned query that reads the condition from the VLM once per chunk, a FiLM-MLP field, and a gripper
head, 13.7M parameters in total and randomly initialized, and integrates $H=8$ steps with RK4
(Appendix~\ref{sec:app_vla}). In ImageWAM, it reads the condition once per chunk
and integrates $H=12$ steps with \textit{Euler} (Appendix~\ref{sec:app_wam}). In GR00T N1.7, a vector field of about
5M parameters, fine-tuned on top of the model, likewise reads the condition once per chunk and integrates $H=16$ steps
with \textit{Euler} (Appendix~\ref{sec:app_setup}).

In RL, it takes one
\textit{Euler} step per control step (Eq.~\eqref{eq:rl_step}), with bounds on the change and on the action. The encoder, the training schedule, and the settings of each baseline are given in
Appendix~\ref{sec:app_setup}.

\textbf{{Evaluation.}} On LIBERO, each visuomotor checkpoint is evaluated with 20 episodes per task from
the fixed initial states of the benchmark, the same for all methods, and we report the success rate
averaged over the four suites. Execution quality is measured on the successful episodes
(Appendix~\ref{sec:app_setup}). Each visuomotor experiment is repeated with three random seeds, and we report
the mean and standard deviation over seeds. On the real robots, each method is trained for 300
epochs (50 on the single-arm robot to test the early performance and 90 on the humanoid) and evaluated with 20
trials. In RL, each actor is trained with three random seeds for 3 hours on one GPU
and evaluated on all 40 clips following SuperMimic.

\section{Simulation Results}
\label{sec:exp_sim}

\subsection{Learning Efficiency}
\label{sec:exp_efficiency}

We first test whether \m{}~policy learns faster, as a visuomotor policy, as the action head of large pretrained
models, and as the actor in RL.

\begin{figure}[!tbp]
\centering
\includegraphics{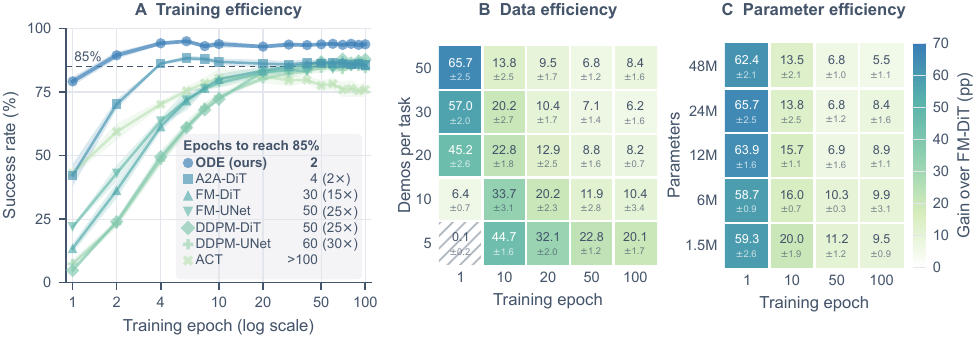}%
\caption{\textbf{Learning efficiency in visuomotor imitation learning.} Success rates on LIBERO, averaged over the
four suites and three random seeds. \textbf{(A)} Success rate during training, with the standard deviation over
seeds as shaded bands. The inset lists the first epoch at which each method reaches 85\% (dashed line) and its ratio
to \m{}~policy. \textbf{(B)} Gain of \m{}~policy over FM-DiT in percentage points with different numbers of
demonstrations per task and \textbf{(C)} of parameters outside the encoder.}
\label{fig:efficiency}
\end{figure}

\subsubsection{Visuomotor imitation learning}

\textbf{{Training efficiency.}} Against the six visuomotor baselines on LIBERO, \m{}~policy learns from the first
epoch (Fig.~\ref{fig:efficiency}(A)). After one epoch, it succeeds in 79.2\% of the episodes, nearly twice the best
baseline, while diffusion and flow policies have barely started. It passes 85\% at epoch 2, an order of magnitude
earlier than flow and diffusion policies, and still leads at epoch 100. Integrating over the physical time of the
robot is thus far easier to learn, without trading off the final performance.

\textbf{{Data efficiency.}} With 5 to 50 demonstrations per task (Fig.~\ref{fig:efficiency}(B)), \m{}~policy leads
FM-DiT at every data scale from epoch 10 on, and the gap grows as data shrink. With 20 demonstrations per task, it reaches 87.7\% at epoch 100, above FM-DiT trained on all
50 (85.4\%). It matches FM-DiT with less than half of the demonstrations.

\textbf{{Parameter efficiency.}} Scaling the parameters outside the encoder from 1.5M to 48M
(Fig.~\ref{fig:efficiency}(C)) barely changes the gain. The smallest \m{}~policy (1.5M) reaches 92.5\% at epoch 10, above the largest FM-DiT (48M) at any
epoch (at most 87.5\%). The advantage comes from the ODE formulation, not from the model capacity.

\begin{figure}[!tbp]
\centering
\includegraphics{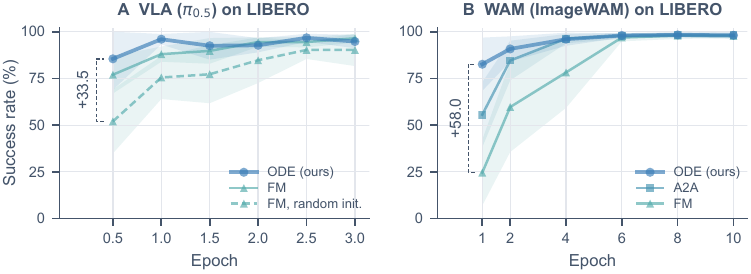}%
\caption{\textbf{Learning efficiency in imitation learning at scale.} Success rate on LIBERO of the ODE head (ours),
the original flow matching head (FM), and an A2A head in (B), one seed. \textbf{(A)} The $\pi_{0.5}$ VLA (10
episodes per task), with the FM expert initialized from $\pi_{0.5}$ (solid) or randomly (dashed). \textbf{(B)}
ImageWAM (25 episodes per task). Brackets mark the gain over the randomly initialized FM expert in (A) and over the
original head in (B), and shaded bands the standard deviation over the four suites. Per-suite results are given in
Appendix~\ref{sec:app_vla} and~\ref{sec:app_wam}.}
\label{fig:scale}
\end{figure}

\subsubsection{Imitation learning at scale}

To test whether the gain carries over to large pretrained models, we replace the action expert of
$\pi_{0.5}$~\citep{intelligence2025pi05} by a randomly initialized ODE head with about 3\% of its parameters
(Fig.~\ref{fig:scale}(A)). Without any pretrained action weights, it leads both FM experts early in training,
with 96.0\% after one epoch against 75.5\% for the randomly initialized one, and the pretrained expert catches up only from the second epoch on. In ImageWAM, which has 4.5B trainable parameters (Fig.~\ref{fig:scale}(B)), the
ODE head succeeds in 82.6\% after one epoch, against 24.6\% for the original head, with A2A~\citep{zuo2026deltaflow}
in between, and all three converge from epoch 6 on. \m{}~policy thus keeps its learning efficiency in large
pretrained models, where the action head is only a small part of the network.

\begin{figure}[!tbp]
\centering
\includegraphics{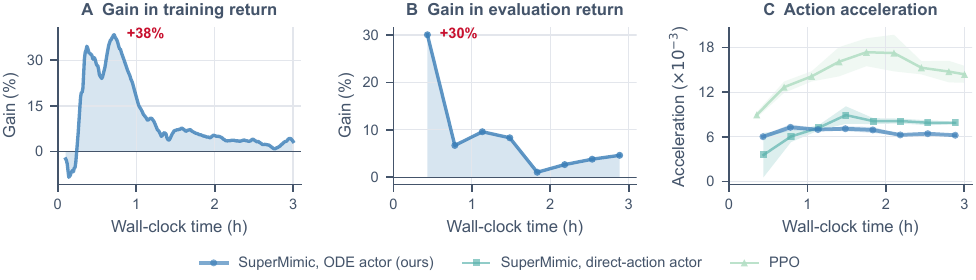}%
\caption{\textbf{Learning efficiency in reinforcement learning.} The ODE actor \textit{vs.}\ the direct-action actor
of SuperMimic and PPO on tracking the 40 LAFAN1 clips with a G1 humanoid, over wall-clock time on one GPU.
\textbf{(A)} Gain of the ODE actor in the training return and \textbf{(B)} in the normalized evaluation return on all
clips, where the steps after a fall count as zero, both from the means over three random seeds. \textbf{(C)} Acceleration of the executed actions (lower is smoother), mean and standard
deviation over three random seeds. At 0.4 hours, the direct-action actor still barely moves.}
\label{fig:rl}
\end{figure}

\subsubsection{Reinforcement learning}
\label{sec:exp_rl}

We finally replace the actor of SuperMimic and keep its world model, critic, rewards, and hyperparameters
(Fig.~\ref{fig:rl}), and train PPO for the same time as a reference without a world model. The ODE actor pulls ahead early. Its training return is 37\% higher after 0.75 hours ($0.630\pm0.007$ against
$0.460\pm0.008$), and its return on all 40 clips is 30\% higher after 0.4 hours and stays ahead until the end
($0.885\pm0.007$ against $0.845\pm0.009$), while PPO reaches only $0.452\pm0.021$ after 3 hours (comparison with
PPO in Fig.~\ref{fig:a_rl_return}). This likely stems from the gradient path along the chain of actions
(Section~\ref{sec:method_rl}). The gradient norm with respect to the first imagined action, relative to the last
one, is about 0.57 for the direct-action actor and 1.0 to 1.1 for the ODE actor, so early actions receive a stronger
learning signal. \m{}~policy further speeds up an already efficient RL method without any change to its algorithm.

\subsection{Execution Quality}
\label{sec:exp_inference}

We next examine how fast \m{}~policy generates actions and how smoothly the robot executes them.

\begin{figure}[!tbp]
\centering
\includegraphics{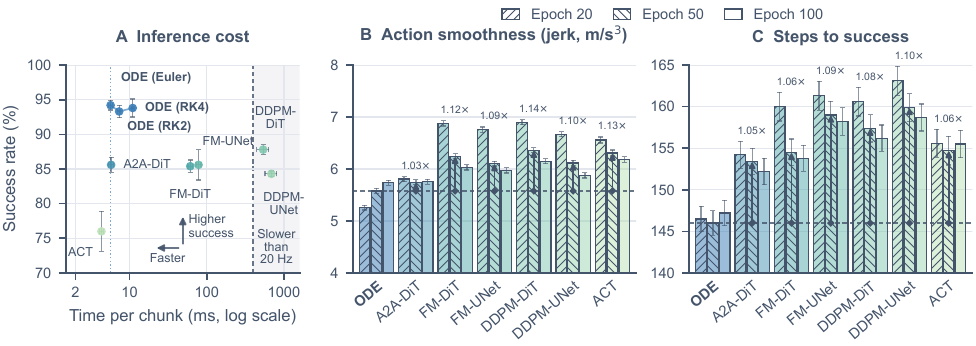}%
\caption{\textbf{Inference cost and action smoothness.} Results on LIBERO. \textbf{(A)} Inference time per chunk of 8
actions \textit{vs.}\ success rate at epoch 100, where the shaded region exceeds the 400\,ms budget of 8 control
steps at 20\,Hz. \m{}~policy is shown with RK4, RK2, and \textit{Euler} (dotted line). Error bars show the standard
deviation of the time over successful episodes of one seed and of the success rate over three random seeds.
\textbf{(B)} RMS jerk of the end-effector position and \textbf{(C)} control steps to complete a task over the
successful episodes at epochs 20, 50, and 100 (hatching), with the standard error over the episodes of three random
seeds. The dashed line marks \m{}~policy at epoch 50, and arrows give the ratio of each baseline to it (lower is
better).}
\label{fig:inference}
\end{figure}

\textbf{{Inference cost.}} Since the solver barely changes the success rate (Fig.~\ref{fig:ablation}(C)), we compare
with \m{}~policy using \textit{Euler} (Fig.~\ref{fig:inference}(A)). It generates a chunk of 8 actions in 5.7\,ms, as
fast as A2A-DiT, 10.8 times faster than FM-DiT, and about 100 times faster than DDPM, which exceeds the budget of a
20\,Hz controller. Only ACT is slightly faster, at a much lower success rate. Even with RK4, it stays several times
faster than FM-DiT, and the ranking holds on an idle GPU (Table~\ref{tab:a_bench}). \m{}~policy thus combines the
speed of single-pass policies with a higher success rate than generative ones.

\begin{figure}[!tbp]
\centering
\includegraphics{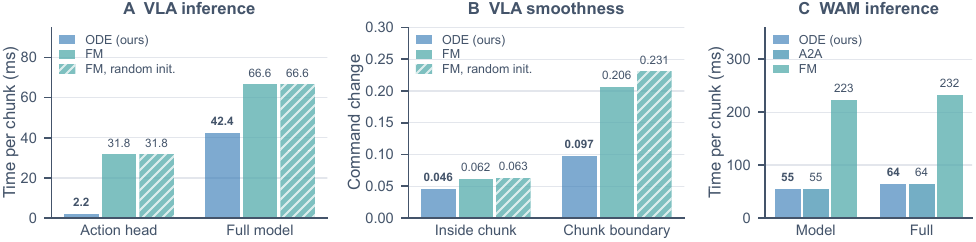}%
\caption{\textbf{Inference cost and action smoothness at scale.} Results on LIBERO with one seed. \textbf{(A)}
Inference time per chunk of 8 actions in $\pi_{0.5}$ at epoch 3, for the action head and for the full model, on one
A100 GPU with batch size 1. \textbf{(B)} Mean change of the normalized position command inside chunks and at the
first step of a new chunk in $\pi_{0.5}$. \textbf{(C)} Inference time per chunk in ImageWAM with cached text
features, for the model and for the full local pipeline. Hatched bars mark randomly initialized action heads.}
\label{fig:scale_exec}
\end{figure}

\textbf{{Inference cost at scale.}} In $\pi_{0.5}$, the ODE head reads the condition once per chunk and runs about 14
times faster than the FM expert with 10 denoising steps, which also saves about a third of the time of the full model
(Fig.~\ref{fig:scale_exec}(A)). Its commands also change about half as much at chunk boundaries
(Fig.~\ref{fig:scale_exec}(B)). In ImageWAM, the action transformer runs once per chunk and a small vector field
carries out the integration, whereas the original head runs the transformer at each of its 10 denoising steps. This
cuts the time per chunk about fourfold, as fast as A2A (Fig.~\ref{fig:scale_exec}(C)). Moreover, each action of
\m{}~policy can be sent to the controller as soon as it is integrated (Section~\ref{sec:method_infer}), whereas the
baselines must first generate the whole chunk.

\textbf{{Action smoothness.}} Over the successful episodes, \m{}~policy has the lowest end-effector jerk at every
checkpoint and finishes tasks in fewer control steps than all baselines (Fig.~\ref{fig:inference}(B, C)). In RL, the
actions of the ODE actor accelerate 21\% less than those of the direct-action actor and less than half as much as PPO after 3 hours (Fig.~\ref{fig:rl}(C)). Since integration links consecutive actions and seamless chunking links the chunks
(Section~\ref{sec:exp_ablation}), the executed motion is both smoother and more efficient.

\subsection{Generalization Performance}
\label{sec:exp_generalization}

We then test how well \m{}~policy generalizes to perturbations unseen in training, as a visuomotor policy and
within a VLA and a WAM.

\begin{figure}[!tbp]
\centering
\includegraphics{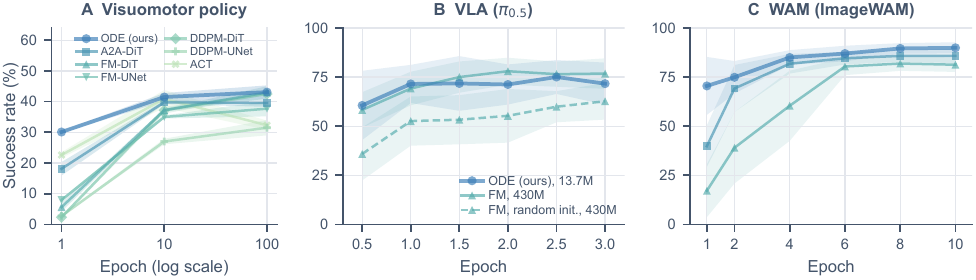}%
\caption{\textbf{Generalization to unseen perturbations.} Zero-shot success rate on LIBERO-plus over all perturbation
types. \textbf{(A)} Visuomotor policies, pooled over the four suites, with the standard deviation over three random
seeds as shaded bands. \textbf{(B)} $\pi_{0.5}$ and \textbf{(C)} ImageWAM on a shared set of 2008 instances, one
seed, with the standard deviation over the four suites as shaded bands. Dashed lines mark randomly initialized action
heads, and the legend of (B) gives the size of each head. Results per perturbation type are given in
Fig.~\ref{fig:plus_types} and Table~\ref{tab:a_wam_plus}.}
\label{fig:plus}
\end{figure}

\subsubsection{Visuomotor imitation learning}
Evaluated zero-shot on LIBERO-plus with 50 perturbed instances per type and suite (Fig.~\ref{fig:plus}(A)),
\m{}~policy leads clearly after one epoch (30.1\% \textit{vs.}\ 22.6\% for ACT) and later stays on par with the
strongest baselines. Per perturbation type (Fig.~\ref{fig:plus_types}), it leads most under sensor noise, FM-DiT and
DDPM-DiT handle a changed initial robot pose better, and a changed camera defeats all methods.

\subsubsection{Imitation learning at scale}
Large pretrained models generalize far better than the visuomotor policies (Fig.~\ref{fig:plus}(B, C)). In $\pi_{0.5}$, the ODE head, with about 3\% of the parameters and no pretrained action
weights, leads the randomly initialized FM expert throughout training. In ImageWAM, \m{}~policy leads throughout training, most clearly early on (70.5\%
\textit{vs.}\ 17.1\% for the original head after one epoch). \m{}~policy reaches its robustness much earlier in training, and on a real robot it also transfers dexterous skills from public data of another lab
(Section~\ref{sec:exp_real_hand}).

\subsection{Multi-Modal Ability}
\label{sec:exp_multimodal}

\begin{figure}[!tbp]
\centering
\includegraphics{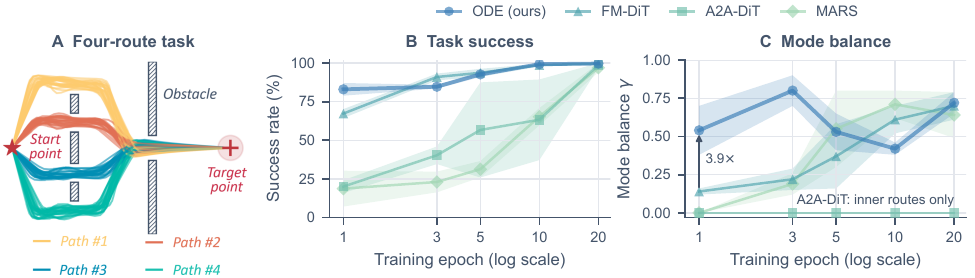}%
\caption{\textbf{Multi-modal tests on a 2D navigation task with four feasible routes.} \textbf{(A)} The agent moves
from the start point through one of four gaps to the target point. \textbf{(B)} Success rate and \textbf{(C)} mode
balance $\gamma$ over training, mean and standard deviation over three random seeds with 100 rollouts each. The
arrow marks the ratio of $\gamma$ between \m{}~policy and FM-DiT after one epoch.}
\label{fig:nav2d_curves}
\end{figure}

Since a single vector field may average distinct behaviors, we test whether \m{}~policy keeps the modes of the
demonstrations on a 2D navigation task, where an agent passes one of four gaps in a wall to reach a goal (50
demonstrations per route) and a policy that averages the routes hits the wall. We report the success rate and the
mode balance $\gamma=4\min_r p_r\in[0,1]$, where $p_r\in[0,1]$ is the share of successful rollouts through gap
$r\in\{1,\dots,4\}$ (Fig.~\ref{fig:nav2d_curves}). Randomizing the direction of the field at the start of a chunk
suffices to send the integrated path into another mode, without a multi-step generative sampling process.

With the diversity head, \m{}~policy takes all four routes from the first epoch (Fig.~\ref{fig:nav2d}). Over the
first five epochs, it is more balanced than FM-DiT, with 3.9 times its $\gamma$ after one epoch, when it also
succeeds more often. FM-DiT is more balanced at epoch 10, and at epoch 20 both are similarly balanced and almost
always succeed. A2A-DiT with initial noise perturbation~\citep{zuo2026deltaflow} learns only the two inner
routes ($\gamma=0$), and MARS covers all routes but learns slowly. \m{}~policy keeps diverse behavior while learning fast, also on a real bimanual task (Section~\ref{sec:exp_real_arm}).

\section{Real-World Results}
\label{sec:exp_real}

We then deploy \m{}~policy on four real robots to test whether its advantages carry over to high-precision,
long-horizon, deformable, multi-modal, dexterous, and loco-manipulation tasks, and finally compare the inference cost and the smoothness across the
robots.

\subsection{Single-Arm Robot}
\label{sec:exp_real_single}

\begin{figure}[!tbp]
\centering
\includegraphics{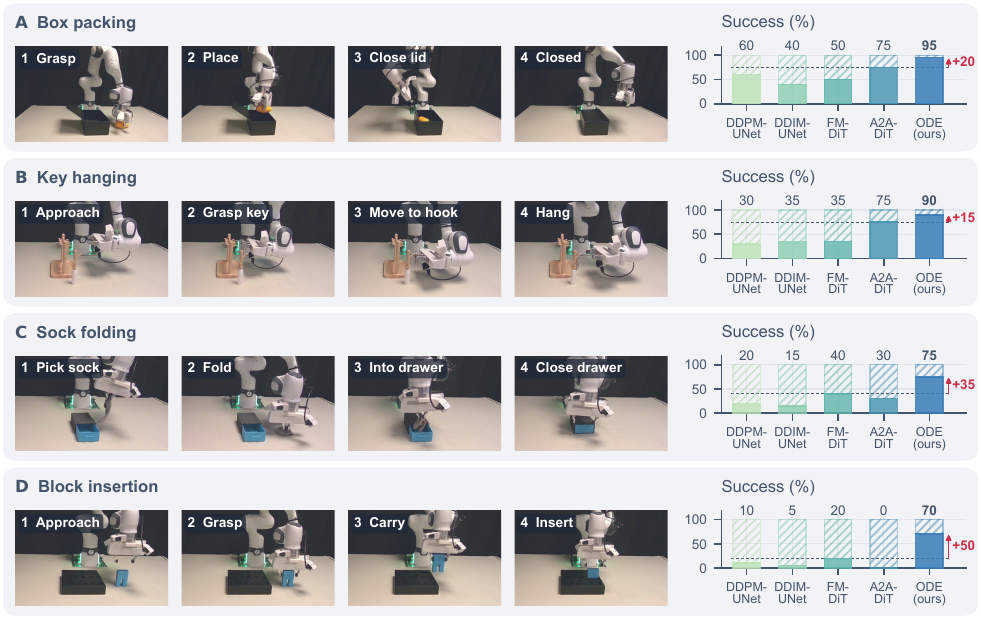}%
\caption{\textbf{Tasks on a real single-arm robot.} \m{}~policy on a Franka Research 3 arm, one row per task.
\textbf{(A)} Putting a mango into a box and closing the lid. \textbf{(B)} Hanging a key on a key holder.
\textbf{(C)} Folding a sock and putting it into a drawer. \textbf{(D)} Inserting a block into a hole of a board. On
the right, success rates over 20 trials, with failed trials hatched and arrows from the best baseline.}
\label{fig:real_franka}
\end{figure}

On a Franka Research 3 arm, all policies are trained on 40 demonstrations per task for only 50 epochs, to test their
performance early in training. \m{}~policy achieves the highest success rate on all four tasks
(Fig.~\ref{fig:real_franka}), with the clearest margins in folding a sock and inserting a block (15 and 14 of 20
trials against 8 and 4 for the best baseline). A2A-DiT comes closest on the first two tasks, likely because it also
generates each chunk from the previous actions, but fails all insertions. At the chunk boundaries, \m{}~policy also commands the lowest acceleration on every task, 3.6, 1.1, 2.5,
and 2.2\,rad/s$^2$, against 2.8 to 9.1 for A2A-DiT and 8.3 to 26.0 for the diffusion and flow matching baselines. It learns multi-stage, deformable, and high-precision tasks reliably from a few dozen demonstrations
within a short training.

\subsection{Dual-Arm Robot}
\label{sec:exp_real_arm}

\begin{figure}[!tbp]
\centering
\includegraphics{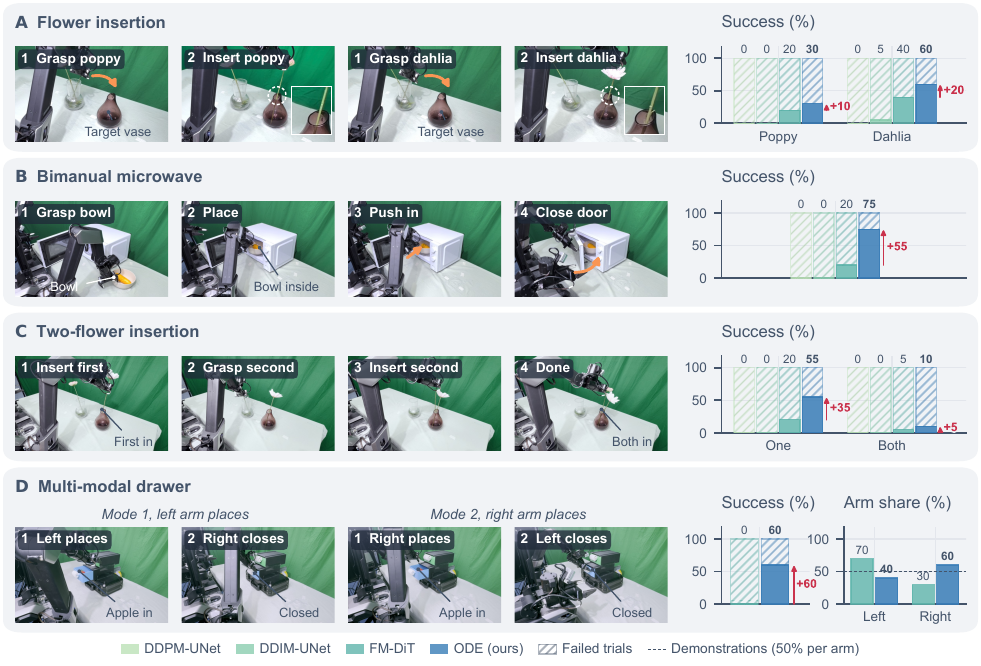}%
\caption{\textbf{Four tasks on a real dual-arm robot.} \m{}~policy on the Galaxea R1 Lite, one row per task.
\textbf{(A)} Inserting a poppy or a dahlia into the narrow opening of a vase (dashed circle and zoom). \textbf{(B)}
Putting a bowl into a microwave, pushing it in, and closing the door. \textbf{(C)} Inserting two flowers one after
the other. \textbf{(D)} Either arm puts an apple into the drawer and the other closes it, two frames per mode. On
the right, success rates over 20 trials per flower or task, with failed trials hatched and arrows from the best
baseline. In (D), the share of trials in which each arm goes for the apple is shown next to it, with the dashed line
at the share in the demonstrations.}
\label{fig:real_all}
\end{figure}

On the Galaxea R1 Lite, \m{}~policy is tested on four tasks, two of them with both arms (Fig.~\ref{fig:real_all}).

\textbf{{High-precision and long-horizon tasks.}} In flower insertion, the thin stem must meet the narrow opening of
a vase, and the long stem of the poppy amplifies small errors at the gripper. \m{}~policy succeeds more often than
FM-DiT with both flowers (60\% \textit{vs.}\ 40\% with the dahlia), likely because its motion stays smooth up to the
opening. In the microwave task, learned from only 34 demonstrations, it succeeds in 75\% of the trials against 20\%
for FM-DiT, which leaves the door open in 9 of its 16 failures, against 2 of 5 for \m{}~policy. Chaining two
insertions into the same vase, it inserts the first flower far more often than FM-DiT (55\% \textit{vs.}\ 20\%),
although the second insertion, into a vase that already holds a stem, is hard for all methods. Both diffusion
policies fail almost every trial of the three tasks. \m{}~policy thus keeps the error of each step small enough for
precise and long sequences with both arms.

\textbf{{Multi-modal task.}} In the drawer task (Fig.~\ref{fig:real_all}(D)), one arm puts an apple into an
open drawer and the other closes it, and the demonstrations assign the two roles both ways equally often. With the
diversity head (Section~\ref{sec:method_side}), \m{}~policy takes both modes, with the left arm placing the apple in 8 of the 20 trials and the right arm in 12, and succeeds in 5 of 8 and
7 of 12, 60\% in total. FM-DiT also goes for the apple with both arms but never
places it. The multi-modality seen in navigation carries over to a bimanual robot.

\subsection{Dexterous Hand}
\label{sec:exp_real_hand}

\begin{figure}[!tbp]
\centering
\includegraphics{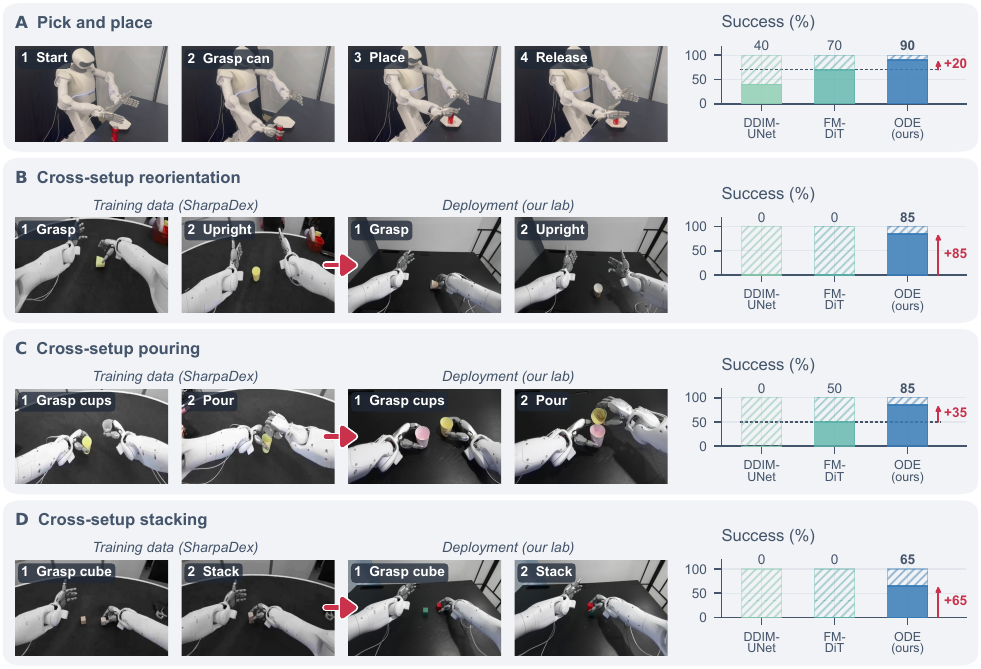}%
\caption{\textbf{Tasks on a real robot with a dexterous hand.} \m{}~policy on the Sharpa North robot with a Sharpa
hand, one row per task. \textbf{(A)} Picking up a can and placing it on a plate. \textbf{(B-D)} Tasks learned only
from public demonstrations of another lab~\citep{sharpadex2026} (left) and deployed in our lab (right),
\textbf{(B)} reorienting a fallen cup upright, \textbf{(C)} pouring pills from one cup into another with two hands,
and \textbf{(D)} stacking one cube on another. On the right, success rates over 20 trials, with failed trials hatched
and arrows from the best baseline.}
\label{fig:real_hand}
\end{figure}

\textbf{{Pick-and-place with a dexterous hand.}} On a Sharpa North robot with a Sharpa dexterous hand, whose joints
greatly enlarge the action space, \m{}~policy picks up a can and places it on a plate in 18 of 20 trials after training on 50
demonstrations, against 14 for FM-DiT and 8 for DDIM-UNet (Fig.~\ref{fig:real_hand}(A)).

\textbf{{Cross-setup transfer.}} We next train on 50 demonstrations per task from the public SharpaDex
dataset~\citep{sharpadex2026}, recorded by Sharpa in its own lab with the same robot and hand, and deploy the
policies in our lab without any demonstration from our setup (Fig.~\ref{fig:real_hand}(B-D)). In reorienting a
fallen cup upright, with other cups and another background, \m{}~policy succeeds in 17 of 20 trials, while FM-DiT
never grasps the cup and DDIM-UNet hovers above it with the hand jittering at the period of one chunk. In pouring
pills between two cups with two hands, using cups of other colors, it succeeds in 17 trials against 10 for FM-DiT and
none for DDIM-UNet. In stacking a red cube on a green one, whereas the demonstrations use wooden cubes, it succeeds
in 13 of 20 trials, while both baselines always fail. \m{}~policy thus carries skills across setups, consistent
with its robustness to unseen perturbations in simulation (Section~\ref{sec:exp_generalization}).

\subsection{Humanoid}
\label{sec:exp_real_humanoid}

\begin{figure}[!tbp]
\centering
\includegraphics{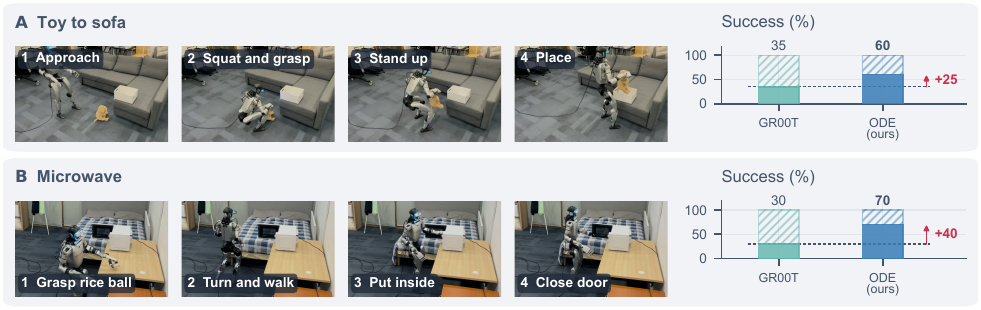}%
\caption{\textbf{Loco-manipulation tasks on a real humanoid.} \m{}~policy as the action head of GR00T N1.7 on a
Unitree G1 with Inspire DFX dexterous hands, one row per task. \textbf{(A)} Walking to a toy on the floor,
squatting to grasp it, standing up, and placing it into a box on the sofa. \textbf{(B)} Grasping a rice ball from a
table, turning and walking to a microwave, putting the rice ball inside, and closing the door. On the right,
success rates over 20 trials, with failed trials hatched and arrows from the baseline.}
\label{fig:real_humanoid}
\end{figure}

We finally test whether \m{}~policy also drives the whole body of a humanoid. On a Unitree G1 with Inspire DFX
dexterous hands, it replaces the action head of GR00T N1.7~\citep{bjorck2025gr00t} (Fig.~\ref{fig:real_humanoid}).
Both tasks are long-horizon and couple walking with a precise grasp. In the toy task, the robot walks to a toy on
the floor, squats to grasp it, stands up, and places it into a box on the sofa, so the legs, the torso, and the
arms must move together. In the microwave task, it grasps a small rice ball from a table, turns and walks to a
microwave, puts the rice ball inside, and closes the door. Trained on 90 demonstrations per task for 90 epochs,
\m{}~policy succeeds in 60\% and 70\% of 20 trials, against 35\% and 30\% for the original head of GR00T.
The vector field thus also carries whole-body loco-manipulation, where small errors of the stance and the grasp
likely accumulate over the long task.

\subsection{Inference Cost and Smoothness}
\label{sec:exp_real_exec}

\begin{figure}[!tbp]
\centering
\includegraphics{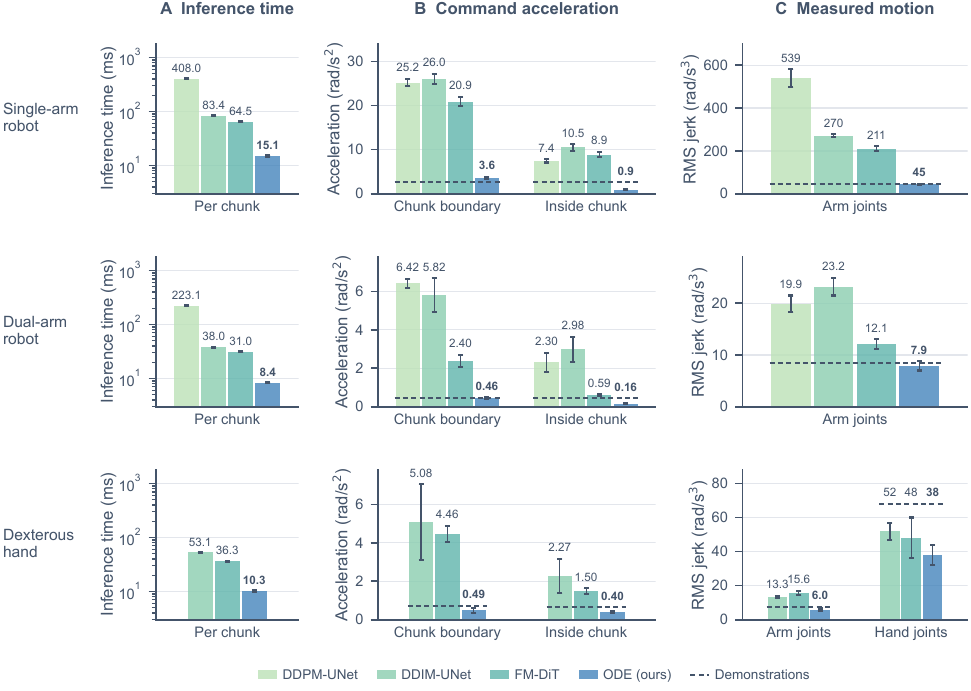}%
\caption{\textbf{Inference cost and smoothness on real robots.} The close-box, microwave, and can tasks (top to
bottom). \textbf{(A)} Inference time per chunk on one RTX 5080, RTX 5090, and RTX 4090, in log scale. \textbf{(B)}
Acceleration of the joint commands at chunk boundaries and inside chunks. \textbf{(C)} RMS jerk of the measured joint
motion. Mean and standard deviation over episodes (successful ones on the single-arm robot), with the demonstrations
as dashed lines.}
\label{fig:real_exec}
\end{figure}

On one task of each of the first three robots, the close-box, microwave, and can tasks
(Fig.~\ref{fig:real_exec}), \m{}~policy generates
a chunk 3.5 to 27 times faster than the diffusion and flow matching baselines, even with RK4. Since each chunk
continues from the end of the previous one, its command acceleration at chunk boundaries stays at or below that of
the demonstrations on the dual-arm robot and the hand and close to it on the single-arm robot, whereas the baselines
reach several times that of the demonstrations. The measured jerk of its arm and hand joints also matches or undercuts the demonstrations, while the arm jerk of the
baselines exceeds them by up to an order of magnitude. A2A-DiT, which also
starts from the previous actions, moves about as smoothly on the single-arm robot (39\,rad/s$^3$) and generates a chunk faster (8.6\,ms), but
accelerates more at chunk boundaries (9.1 against 3.6\,rad/s$^2$). Appendix~\ref{sec:app_real} reports the same measures on the cup and pill
tasks.

\begin{figure}[!tbp]
\centering
\includegraphics{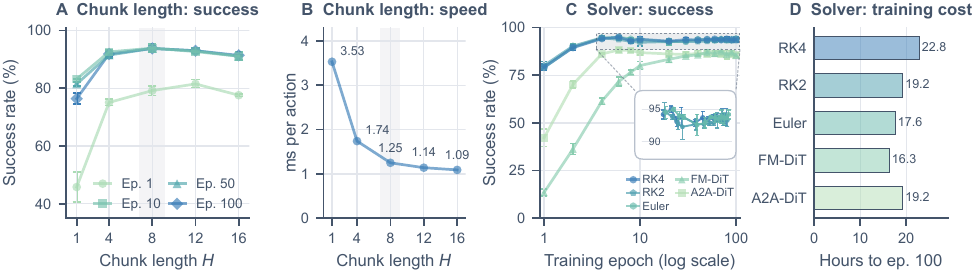}%
\caption{\textbf{Ablations of the chunk length and the integration solver.} Results of \m{}~policy on LIBERO,
averaged over the four suites. \textbf{(A)} Success rate with chunk length $H$ at epochs 1, 10, 50, and 100.
\textbf{(B)} Inference time per action at epoch 100 on one idle GPU, with the default $H=8$ in gray. \textbf{(C)}
Success rate during training with three solvers, with FM-DiT and A2A-DiT as references and an inset on the solvers.
\textbf{(D)} Training time to epoch 100 (one RTX 4090 per run, eight runs in parallel). Mean and standard deviation
over three random seeds (two for $H=16$, as training with the third diverged).}
\label{fig:ablation}
\end{figure}

\section{Ablation Studies}
\label{sec:exp_ablation}

\begin{figure}[!tbp]
\centering
\includegraphics{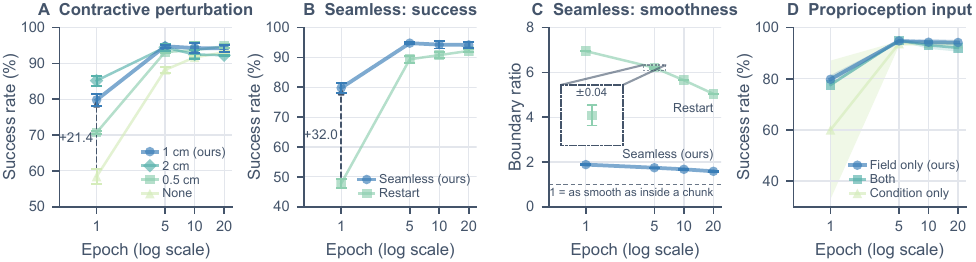}%
\caption{\textbf{Ablations of the contractive perturbation, seamless chunking, and proprioception.} Results of
\m{}~policy on LIBERO, averaged over the four suites (10 episodes per task). \textbf{(A)} Without the contractive
perturbation and with three magnitudes of it. \textbf{(B, C)} Seamless chunking \textit{vs.}\ restarting each chunk
from the measured state, success rate and boundary ratio (command change at the first step of a chunk divided by
that inside chunks). \textbf{(D)} Proprioception fed to the vector field, to both the condition and the field, or to
the condition. Arrows in (A) and (B) mark the gap after one epoch. Mean and standard deviation over three random
seeds, with an inset in (C) on the largest standard deviation.}
\label{fig:ablation2}
\end{figure}

We ablate the design choices of \m{}~policy one at a time with all other settings fixed, namely the chunk length and
the solver (Fig.~\ref{fig:ablation}), the contractive perturbation, seamless chunking, and the placement of the
proprioception (Fig.~\ref{fig:ablation2}), and the number of components of the diversity head
(Fig.~\ref{fig:ablation_k}). The diversity head at inference and the architecture of the vector field are ablated in
Appendix~\ref{sec:app_ablation}.

\textbf{{Chunk length.}} With chunk length $H=1$ in training and execution (Fig.~\ref{fig:ablation}(A, B)), the
policy reaches only 76.4\% at epoch 100 and is the slowest per action, since the encoder runs at every step. From
$H=4$ on, the success rate stays above 91\% (best at the default $H=8$), while the time per action keeps falling,
although training shows occasional loss spikes for $H\ge12$. Consistent with the finding that chunking mitigates the
compounding errors of behavioral cloning~\citep{lazzati2026,park2026bcmystery}, chunking is essential, and $H=8$
balances success rate and inference time.

\textbf{{Integration solver.}} Replacing RK4 with RK2 or \textit{Euler} changes the success rate at epoch 100 by
less than one point, within the variation across seeds, and all three stay above FM-DiT and A2A-DiT throughout
training (Fig.~\ref{fig:ablation}(C)). The cheaper solvers shorten both training and inference
(Fig.~\ref{fig:ablation}(D)). With \textit{Euler}, training takes about as long as for FM-DiT and less than for
A2A-DiT, so the gain does not come from a longer training. Since even \textit{Euler} applies the same vector field at
every step, the performance comes from the ODE formulation, not from an accurate solver.

\textbf{{Contractive perturbation.}} Without the contractive perturbation (Fig.~\ref{fig:ablation2}(A)), the success rate after one epoch drops from 79.8\% to 58.4\% (with 10 episodes per task, hence 79.8\% instead of
79.2\% for the default), and the gap closes by epoch 20. Early in training, the inaccurate
field lets seamless chunking drift away from the demonstrations, and the perturbation teaches it to return. A larger
magnitude learns faster, but twice the default later stays slightly lower, likely because large offsets pull the
field away from the fine motion of the demonstrations. Even without the perturbation, \m{}~policy is still well
ahead of FM-DiT and A2A-DiT after one epoch.

\textbf{{Seamless chunking.}} Restarting each chunk from the measured state instead (Fig.~\ref{fig:ablation2}(B, C))
discards the lead of the desired action over the measured state and lowers the success rate after one epoch from
79.8\% to 47.8\%. After training, the success rates are close, but the restarted policy jumps at every chunk
boundary, with a 3 to 4 times higher boundary ratio, a 1.3 to 2.2 times higher jerk, and 8\% to 19\% longer episodes. Seamless chunking is thus
key to both fast learning and smooth execution.

\textbf{{Proprioception.}} Feeding the proprioception to the vector field only, to both the condition and the field,
or to the condition only performs similarly from epoch 5 on (Fig.~\ref{fig:ablation2}(D)). After one epoch, the
field-only variant is the highest, while the condition-only variant varies strongly across seeds. Feeding it to the field performs at least as well as the conventional placement and makes early learning more reliable.

\begin{figure}[!tbp]
\centering
\includegraphics{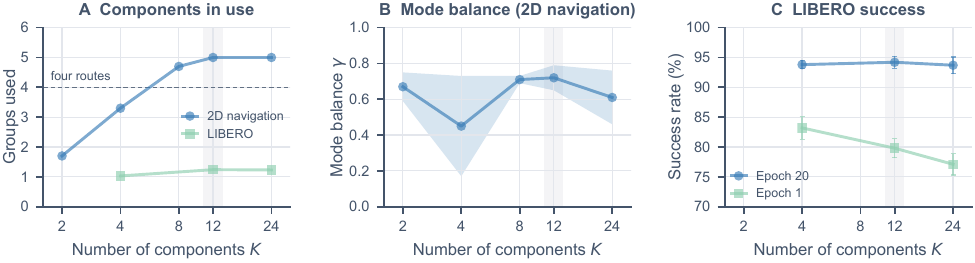}%
\caption{\textbf{Ablation on the number of components.} \textbf{(A)} Distinct groups of components used by the
diversity head at the start of the 2D navigation task and per state on LIBERO, with the four routes as dashed line.
\textbf{(B)} Mode balance $\gamma$ on the navigation task at epoch 20. \textbf{(C)} Success rate on LIBERO at epochs 1
and 20 (10 episodes per task). Mean and standard deviation over three random seeds, with the default $K=12$ in
gray.}
\label{fig:ablation_k}
\end{figure}

\textbf{{Number of components.}} The number of components $K$ only bounds how many modes the diversity head can
represent (Fig.~\ref{fig:ablation_k}). At $K=4$, the head cannot reliably separate the four routes of the navigation task, and $\gamma$
varies strongly across seeds. From $K=8$ on, it uses about five groups of components at the start,
close to the four routes, and about one group per state on uni-modal LIBERO. A smaller $K$ is slightly better after
one epoch, likely because the untrained head offsets the actions less, and at epoch 20 the success rate barely
changes. $K$ only needs to exceed the number of modes, and we use $K=12$ throughout.

\section{Discussion and Future Directions}
\label{sec:discussion}

We presented \m{}~policy, which models how the action of a robot evolves over its physical time as an ODE,
so that each action is integrated from the previous one. In IL, it learns the vector field by
integrating over short action chunks, seamless chunking joins successive chunks without jumps, and the
contractive perturbation keeps the policy close to the demonstrations. On LIBERO, \m{}~policy reaches 79.2\% after one epoch and 93.8\% at
epoch 100, matches FM-DiT with less than half of the demonstrations, and generates a chunk of 8 actions in
5.7\,ms. As the action head of the $\pi_{0.5}$ VLA, it reaches 96.0\% after one
epoch against 75.5\% for its flow matching expert with about 3\% of its parameters, and as the action head
of the WAM ImageWAM, 82.6\% against 24.6\% for the original head, generating a chunk 4.1 times faster.
On 14 real-world tasks across single-arm, dual-arm, dexterous, and humanoid robots, it achieves the highest success
rate on every task, and its motion is smoother than the human demonstrations in most cases.
As the actor of SuperMimic, it reaches a higher return early in training and still 0.885 against 0.845 at the end
of training.

These results suggest that building the continuity of physical motion into the policy, rather than learning
it from data, makes robot policies more efficient in learning, data, parameters, and execution, and that this gain scales from
small visuomotor policies to large pretrained models and carries over from IL to RL. Beyond the results, viewing a policy as a vector field in the physical time of the robot opens uses
that direct-action policies do not naturally offer. We share three aspects with preliminary experiments, as
starting points rather than finished solutions, and close with a broader view of what the field may represent.

\begin{figure}[!tbp]
\centering
\includegraphics{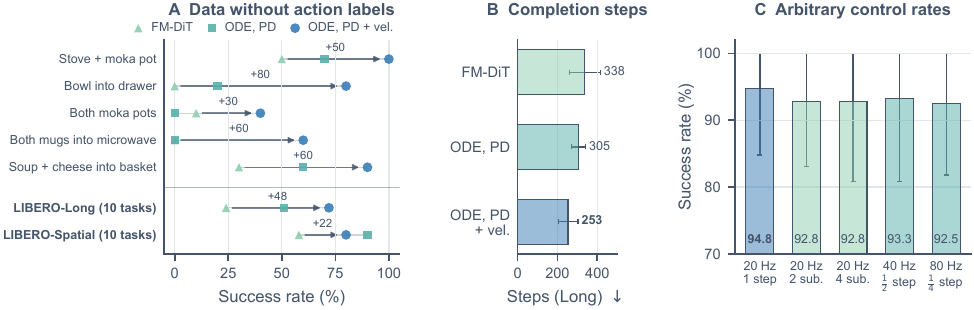}%
\caption{\textbf{Two potential uses of the continuous-time field.} \textbf{(A, B)} Learning from data without action labels. All policies are trained on the measured poses of LIBERO only and evaluated at epoch 100, and \m{}~policy with
PD $+$ velocity feeds the velocity of its field to the PD controller. (A) Success rate on five time-critical
LIBERO-Long tasks and averaged over LIBERO-Long and LIBERO-Spatial, and (B) steps of successful Long episodes. \textbf{(C)} Arbitrary control rates, by changing the integration of one trained
model at deployment, with substeps within each control step at 20\,Hz or a smaller step sending every
intermediate point at 40 or 80\,Hz. Average success rate over the four suites. One seed with 10 episodes
per task. In (A), arrows give the gain of \m{}~policy with PD $+$ velocity over FM-DiT. Whiskers in (B) and
(C) show the standard deviation across tasks.}
\label{fig:discussion}
\end{figure}

\subsection{Learning from Data without Action Labels} Data collected with handheld interfaces such
as UMI~\citep{chi2024universal} or from human videos~\citep{kareer2025emergence} record the measured
pose but not the command sent to the controller. A policy can then only learn the measured state,
which lags behind its target, so the robot moves slower than in the demonstrations~\citep{zhao2026apex}. Since
\m{}~policy learns a vector field, the velocity of each action is available at no extra cost. We feed
it to the damping term of the PD controller, i.e.,
$k_p(\mathbf{p}_{\text{ref}}-\mathbf{p})+k_d(\dot{\mathbf{p}}_{\text{ref}}-\dot{\mathbf{p}})$
instead of $k_p(\mathbf{p}_{\text{ref}}-\mathbf{p})-k_d\dot{\mathbf{p}}$, where
$\mathbf{p},\mathbf{p}_{\text{ref}}\in\mathbb{R}^{d_a}$ are the measured and reference poses and
$k_p,k_d\in\mathbb{R}_{>0}$ are the proportional and derivative gains, and train all policies on
the measured poses of LIBERO only (Fig.~\ref{fig:discussion}(A, B)). With the velocity, the robot keeps the demonstrated pace. Successful Long episodes take 253 instead of 305 steps, and long tasks that
ran out of time are now completed, raising Long from 51\% to 72\% (FM-DiT: 24\%). Faster motion,
however, makes precise placement harder, and on Spatial the success rate drops from 90\% to 80\%. This
hints that the velocity could be modulated by the phase of a task, e.g., fast in free space and slow near
contact, which is natural to express with a vector field but hard with direct-action policies.

\subsection{Arbitrary Control Rates} Since \m{}~policy is a vector field in continuous time, the
integration step is a choice made at deployment rather than a part of the model. We take the model
trained with one RK4 step per control step at 20\,Hz and change only the integration at deployment
(Fig.~\ref{fig:discussion}(C)). Integrating each control step with 2 or 4 substeps moves the integrated
path by at most 1\,mm, against about 10\,mm between consecutive actions, and the success rate changes
from 94.8\% to 92.8\%. Reducing the step to $1/2$ or $1/4$ and sending every intermediate point runs
the same model at 40 or 80\,Hz, with 16 or 32 actions per chunk, at 93.3\% and 92.5\% success and
the same completion time. In contrast, a conventional policy outputs actions at the fixed step of its
training data and must interpolate them to run at a higher rate, and even policies conditioned on physical
time keep a fixed action rate~\citep{wang2026chronos}. One model could therefore serve
low-level controllers of different frequencies without retraining, and the integration step might even be
adapted online, e.g., refined near contact where precise control matters most.

\subsection{Discrete Action Signals}
\label{sec:disc_signals}
An ODE describes signals that change continuously. Contact tasks nevertheless do not hurt \m{}~policy, e.g., 96.2\%
\textit{vs.}\ 78.8\% for FM-DiT at epoch 10 on eight contact-rich LIBERO tasks. The gripper command, in contrast, is
binary on some robots, e.g., in LIBERO, while it is continuous on our real robots. The LIBERO experiments above
predict the binary command with a separate classification head. This simple head is effective and leaves the
smooth motion of the arm to the field. To probe the ODE itself, we instead integrate the
gripper as one more state of the field and send its sign, with each chunk starting the gripper state at
the last command. Without any further
design, the success rate drops from 79.2\%, 95.0\%, and 92.8\% to 74.5\%, 91.8\%, and 92.2\% at epochs 1, 5, and
10, mostly on Long, which requires repeated grasping and releasing. Even so, it still exceeds all baselines after
one epoch (74.5\% \textit{vs.}\ at most 43.2\%) and at epoch 10 (92.2\% \textit{vs.}\ at most 86.9\%).
Neural ODEs built for discontinuous dynamics, such as learned event functions~\citep{chen2021event} and jump
terms~\citep{jia2019jump}, might further empower ODE policy.

Actions learned without demonstrations need not be continuous either. In our early experiments with model-free RL, the learned actions often jumped between control steps, while the compliant PD controller still kept
the motion smooth. There, \m{}~policy brought little gain and sometimes even hurt, likely because its continuity
also smooths the exploration. Such jumps are, however, usually treated as a flaw rather than a need, since they
load the motors and widen the gap between simulation and reality, and RL for robots commonly penalizes the change of
actions~\citep{mysore2021caps,rudin2022learning}. \m{}~policy thus suits best where the commands themselves should be
smooth, e.g., when imitating demonstrations or deploying on hardware.

\subsection{Actions as Unfolding Intentions}
\label{sec:disc_intention}
In physics, differential equations describe how a state changes. The action of a robot is different. It is a
command, an intention of where to go next, so it is not obvious that it should follow such a law. Neuroscience
suggests that it may. When a monkey reaches, its motor cortex does not simply encode the target. Its activity
evolves like a dynamical system from a preparatory state, and the movement unfolds from
it~\citep{churchland2012neural,shenoy2013cortical}. Human reaching is also smooth, close to minimum
jerk~\citep{flash1985coordination}. \m{}~policy works in a similar way. At the start of each chunk, it makes a
discrete choice, e.g., which arm places the apple, and the vector field then unfolds the motion. A policy might
thus only need to decide at a few key points of a task, e.g., where its modes branch, and leave the rest to the
learned dynamics.

\section*{Acknowledgments}
We thank Sharpa Robotics for releasing the SharpaDex dataset~\citep{sharpadex2026}, whose demonstrations we use for the cup reorientation, pill pouring, and cube stacking tasks, and for their support with the Sharpa North robot and the Sharpa hand.

\bibliographystyle{assets/plainnat}
\bibliography{bib/paper}

\clearpage
\newpage
\appendix
\setcounter{figure}{0}\setcounter{table}{0}
\renewcommand{\thefigure}{S\arabic{figure}}\renewcommand{\thetable}{S\arabic{table}}
\renewcommand{\theHfigure}{S\arabic{figure}}\renewcommand{\theHtable}{S\arabic{table}}
\newcommand{\appsec}[2]{\par\noindent\hyperref[#1]{\textbf{\ref*{#1}\quad #2}}\hfill\textbf{\pageref*{#1}}\par\vspace{2pt}}
\newcommand{\appsub}[2]{\par\noindent\hspace{1.5em}\hyperref[#1]{\ref*{#1}\quad #2}\ \dotfill\ \pageref*{#1}\par}
\noindent{\Large\sffamily\bfseries\color{marshead}Contents of the Appendix}\par\vspace{8pt}
{\small
\appsec{sec:app_setup}{Experimental Details}\vspace{4pt}
\appsec{sec:app_results}{Additional Results}
\appsub{sec:app_execution}{Execution Quality}
\appsub{sec:app_generalization}{Generalization}
\appsub{sec:app_multimodal}{Multi-Modality}
\appsub{sec:app_vla}{Vision-Language-Action Model}
\appsub{sec:app_wam}{World Action Model}
\appsub{sec:app_real}{Real-World Experiments}
\appsub{sec:app_rl}{Reinforcement Learning}
\appsub{sec:app_ablation}{Ablation Studies}
}
\vspace{12pt}

\section{Experimental Details}
\label{sec:app_setup}

\textbf{{Simulated benchmarks.}} LIBERO has four suites (Object, Spatial, Goal, and Long) with 10 tasks each and 50
demonstrations per task. One model is trained on all 40 tasks and evaluated on each suite separately.
LIBERO-plus perturbs the camera, the initial robot pose, the object layout, the background, the lighting, and the
sensor noise. $\pi_{0.5}$ and ImageWAM are tested on LIBERO and on a shared set of 2008 LIBERO-plus instances, which also perturb the language
instruction.

\textbf{{Single-arm robot tasks.}} All four tasks on the Franka Research 3 arm use multiple cameras, and the actions
are joint angles. In the box task, the arm puts a mango into a box and then closes the lid. Its
40 demonstrations are collected with a SpaceMouse through the HIL-SERL teleoperation
framework~\citep{luo2025precise}. In the key task, the arm grasps a key and hangs it on a hook of a wooden key
holder, learned from 40 demonstrations collected in the same way. In the sock task, the arm takes a sock from the
top of a small cabinet, folds it on the table, puts it into the open drawer, and closes the drawer, learned from
40 demonstrations. In the insertion task, the arm grasps a blue block and inserts it into a matching hole of a black
board, learned from 40 demonstrations. Inference times on this robot are
measured on one RTX 5080.

\textbf{{Dual-arm robot tasks.}} All tasks on the dual-arm and dexterous-hand robots use joint angles as actions at
15\,Hz. In the flower insertion task, one arm of the Galaxea R1 Lite grasps a long-stemmed poppy or a short-stemmed
dahlia from a glass vase and inserts its thin stem into the narrow opening of another vase, so a small error of
the gripper pose makes the stem miss the opening. One model is trained on both flowers from 95 demonstrations
(38 poppy and 57 dahlia), with a head and a wrist camera. In the microwave task, one arm grasps a bowl of corn,
places it in the microwave, and pushes it in, and the other arm then closes the door. It is learned from 34
demonstrations, with a head and two wrist cameras. In the two-flower task, one arm inserts two flowers one after the other into the
same vase, learned from 95 demonstrations. In the drawer task, one arm puts an apple from the
top of a cabinet into its open drawer, and the other arm closes the drawer. The demonstrations contain both
assignments of the arms, 25 with the left arm and 25 with the right arm placing the apple, with a head and two wrist cameras.

\textbf{{Dexterous-hand tasks.}} The policy runs at 15\,Hz, and each predicted chunk is linearly interpolated to
60\,Hz joint position targets for the arm and the 22 joints of each hand. In the can task, the Sharpa North robot
picks up a can with its Sharpa hand and places it on a plate, learned from 50 demonstrations. In the cup task, the hand picks up a fallen cup,
reorients it, and places it upright. Its 50 demonstrations are drawn at random from the reorient\_cup\_upright task of
the public SharpaDex dataset~\citep{sharpadex2026}, recorded by Sharpa in another lab, and the policies are tested in
our lab with other cups and another background. In the pill task, the two hands pour pills from one cup into
another and put both cups back. Its 50 demonstrations are those of one recording session of the
transfer\_pills\_between\_cups task of SharpaDex closest to our placement of the cups, and the policies are tested
with cups of other colors. In the stacking task, the hand stacks a red cube on a green cube. Its 50 demonstrations
come from the stack\_cube task of SharpaDex, and the policies are tested in our lab with colored cubes instead of the wooden cubes
of the demonstrations.

\textbf{{Humanoid tasks.}} The Unitree G1 humanoid with two Inspire DFX dexterous hands is controlled through
GR00T N1.7~\citep{bjorck2025gr00t}. As in ImageWAM, \m{}~policy reads the condition from GR00T once per chunk.
A vector field of about 5M parameters, fine-tuned on top of GR00T, integrates chunks of $H=16$ steps with
\textit{Euler}, with the diversity head as in all IL experiments. The baseline is the original action head of
GR00T N1.7.

\textbf{{Visuomotor policies.}} All methods share the same observation encoder, a ResNet-18~\citep{he2016deep}
with group normalization~\citep{wu2018group} and random cropping that encodes the last 8 image frames, and a CLIP text
embedding~\citep{radford2021clip} of the instruction. Except in the
model-size study, the parameters outside the encoder are matched to about 24M (ODE-MLP 24.40M, ODE-DiT
24.15M, FM-DiT 23.93M, and the others within 5\%). All methods are trained with AdamW~\citep{loshchilov2019adamw},
learning rate $10^{-4}$, batch size 32, and a cosine schedule defined over 100 epochs, and shorter runs stop
early under the same schedule. The model-size study changes the width of the action head, with
4 layers for both methods. For the 1.5M \m{}~policy, the network that maps the observation features to the
condition is also narrowed (condition width 136 instead of 512), since it alone has 4.2M parameters at the default
width. Each visuomotor experiment is run with three random seeds, except ODE-DiT with one seed. The data-efficiency study changes only the number of demonstrations per task,
so fewer demonstrations also mean fewer gradient steps per epoch.

\textbf{{\m{}~policy.}} The vector field is a 4-layer FiLM-MLP of width 1732 that takes the action state and the proprioception as input and is modulated by the condition. It corresponds to the conditional residual block of the 1D UNet of FM-UNet and DDPM-UNet without the temporal convolution, since the field takes a single state rather than a sequence of actions, and we therefore call it ODE-MLP in the tables. ODE-DiT replaces it with AdaLN Transformer
blocks of width 512 and 4 layers (Appendix~\ref{sec:app_ablation}). The state is the desired action of the LIBERO operational space controller, i.e., the current pose plus 0.05\,m and 0.5\,rad times the normalized command,
so that a unit tracking gain reproduces the demonstrated command exactly. Each chunk of 8 steps is
integrated with RK4 at 20\,Hz. Half of the training segments are perturbed by about 1\,cm and 0.03\,rad with
$\rho=0.7$. The diversity head has $K=12$ components. LIBERO uses the component means and the navigation task
samples them. The binary gripper is predicted by a separate head.

\textbf{{Baselines.}} FM-DiT, FM-UNet, DDPM-DiT, and DDPM-UNet predict chunks of 8 actions with the DiT or the
1D UNet head of the corresponding generative model. A2A-DiT starts its flow from the history of actions,
perturbed by \textit{Gaussian} noise of standard deviation 0.5. ACT uses its CVAE Transformer decoder.
FM-DiT and FM-UNet integrate their flow with 10 midpoint steps, i.e., 20 network calls per chunk, DDPM-DiT and
DDPM-UNet denoise in 100 steps, and A2A-DiT generates a chunk in one flow step. As for \m{}~policy, every
baseline reads the last 8 image frames, predicts a chunk of 8 actions, and executes all of them before the next
observation.

\textbf{{Evaluation and metrics.}} Each LIBERO checkpoint runs 20 episodes for each of the 10 tasks of each
suite from the fixed initial states of the benchmark. Execution quality is measured on the successful
episodes. The jerk is the RMS jerk of the measured end-effector position. The in-chunk and cross-chunk
smoothness are the mean change of the command between consecutive steps inside a chunk and at the first
step of a new chunk, and their quotient is the boundary ratio. The completion time is the number of control
steps. Inference times in the main text are measured during these evaluations on shared GPUs, and times on one idle RTX 4090 are reported in Table~\ref{tab:a_bench}. LIBERO-plus samples 50 instances per perturbation type
and suite with the same sampling seed for all methods.

\section{Additional Results}
\label{sec:app_results}

\subsection{Execution Quality}
\label{sec:app_execution}

Table~\ref{tab:a_bench} repeats the timing of Fig.~\ref{fig:inference}(A) on one idle RTX 4090, and the ranking
stays the same. With
RK4, \m{}~policy takes 8.82\,ms per call, 4.8 times less than FM-DiT and 38 times less than DDPM-UNet, and only
the single-pass ACT and A2A-DiT are faster. With \textit{Euler}, it takes 4.25\,ms, close to these single-pass
policies. The encoder takes about 1.2\,ms for every method, so the differences come from the action head.
Longer chunks spread one call over more actions, and the time per action falls from 3.53\,ms at $H=1$ to
1.09\,ms at $H=16$.

\begin{table}[t]
\centering
\caption{Inference time (ms) on one idle RTX 4090 with the epoch-100 checkpoints. Micro timings repeat a fixed input 200 times, and call and action timings are measured in closed loop on the first three tasks of LIBERO-Goal (10 episodes each), where SR is the success rate (\%). Params counts all parameters (M).}
\label{tab:a_bench}
\footnotesize
\begin{tabular}{lccccccc}
\toprule
Method & Micro call & Micro encoder & Micro head & Call & Action & SR & Params \\
\midrule
ODE-MLP (ours) & 8.82 & \underline{1.22} & 7.6 & 9.69 & 1.25 & \underline{96.7} & 35.6 \\
FM-DiT & 42.1 & 1.28 & 40.8 & 41.3 & 5.31 & 93.3 & 35.1 \\
FM-UNet & 58.8 & \underline{1.22} & 57.5 & 60.1 & 7.7 & 86.7 & 36.0 \\
A2A-DiT & 3.68 & 1.24 & 2.44 & 4.79 & \underline{0.617} & \underline{96.7} & 35.1 \\
DDPM-DiT & 235 & 1.24 & 233 & 237 & 30.5 & 90.0 & 34.7 \\
DDPM-UNet & 335 & 1.25 & 333 & 341 & 43.5 & 93.3 & 36.0 \\
ACT & \textbf{3.19} & 1.26 & \textbf{1.94} & \underline{4.29} & \textbf{0.549} & 80.0 & 35.2 \\
\midrule
ODE-MLP, $H=1$ & \underline{3.39} & \textbf{1.21} & \underline{2.18} & \textbf{3.53} & 3.53 & 80.0 & 35.6 \\
ODE-MLP, $H=4$ & 6.22 & 1.28 & 4.94 & 6.86 & 1.74 & \textbf{100.0} & 35.6 \\
ODE-MLP, $H=12$ & 12.4 & 1.24 & 11.2 & 13.1 & 1.14 & \underline{96.7} & 35.6 \\
ODE-MLP, $H=16$ & 15.8 & \underline{1.22} & 14.5 & 16.2 & 1.09 & \textbf{100.0} & 35.6 \\
\midrule
ODE-MLP, RK2 & 5.65 & \textbf{1.21} & 4.44 & 6.33 & 0.819 & \textbf{100.0} & 35.6 \\
ODE-MLP, \textit{Euler} & 4.25 & 1.23 & 3.02 & 4.86 & 0.623 & \textbf{100.0} & 35.6 \\
\bottomrule
\end{tabular}
\par\vspace{2pt}{\scriptsize Unless noted, ODE-MLP uses $H=8$ and RK4.}
\end{table}

\subsection{Generalization}
\label{sec:app_generalization}

\begin{figure}[!tbp]
\centering
\includegraphics{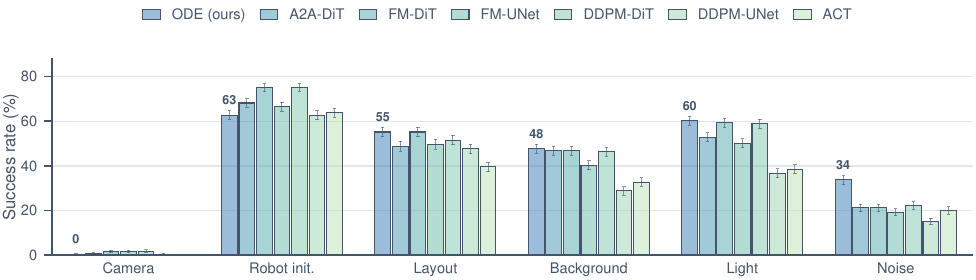}%
\caption{Zero-shot success rate of the visuomotor policies per LIBERO-plus perturbation type at epoch 100, pooled over
the four suites and three random seeds. The number above each group is \m{}~policy, and error bars show the
standard error over the pooled instances.}
\label{fig:plus_types}
\end{figure}

Fig.~\ref{fig:plus_types} breaks the zero-shot success of the visuomotor policies on LIBERO-plus down by perturbation
type at epoch 100, as summarized in Section~\ref{sec:exp_generalization}.

\subsection{Multi-Modality}
\label{sec:app_multimodal}

\begin{figure}[!tbp]
\centering
\includegraphics[width=\linewidth]{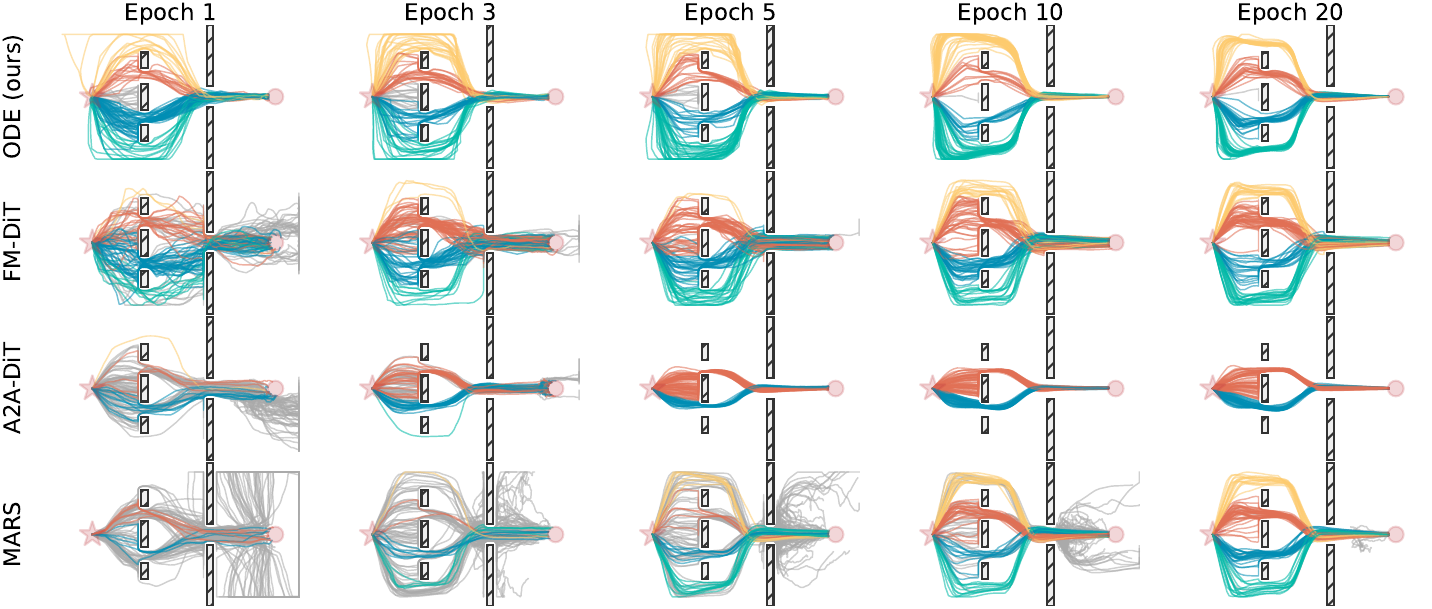}
\caption{Rollouts on the 2D navigation task at epochs 1, 3, 5, 10, and 20, 100 per panel (stars mark the start, circles the goal, and colors the gap taken, with failed rollouts in gray). One of the three random seeds.}
\label{fig:nav2d}
\end{figure}

Fig.~\ref{fig:nav2d} shows the rollouts behind Fig.~\ref{fig:nav2d_curves} for one seed. After one epoch,
\m{}~policy already passes through all four gaps with few failures, whereas FM-DiT fails in many rollouts and
mostly takes the inner routes. A2A-DiT collapses to the two inner routes throughout training. MARS covers all four
routes, but many of its rollouts still fail at epoch 10.

\subsection{Vision-Language-Action Model}
\label{sec:app_vla}

We fine-tune $\pi_{0.5}$~\citep{intelligence2025pi05} on the four LIBERO suites jointly (1693 demonstrations) for
three epochs with one seed. All parameters are trained, including the VLM of 2.92B parameters, with AdamW, a peak
learning rate of $5\times10^{-5}$, and a global batch size of 384 on eight A100 GPUs. Each method uses the current
images of an external and a wrist camera at $224\times224$ and predicts and executes chunks of 8 actions. \m{}~policy
replaces the action expert of $\pi_{0.5}$ (430M parameters with its projections) by a learned query that reads the image and language
features of the VLM once per chunk, a FiLM-MLP vector field of four residual layers of width 512, and a gripper
head, 13.7M parameters in total, all randomly initialized. As in all IL experiments, it also uses the diversity
head. The proprioception enters only the field and the
gripper head. The field integrates each chunk with RK4 from the end of the previous one, with the contractive
perturbation (1\,cm and 0.03\,rad, applied to half of the training chunks). FM integrates its flow with 10
\textit{Euler} steps. The VLM is pretrained in all methods, and the suffix -R only marks a randomly initialized action
expert. Each LIBERO checkpoint is
evaluated on the 10 fixed initial states of each task, and LIBERO-plus on the shared set of 2008 instances with one
episode each. The parameters of \m{}~policy outside the VLM are about 3\% of those of the experts, so the
comparison is not at matched capacity.

Table~\ref{tab:a_vla} lists the success rate per suite at epoch 1. The early lead of \m{}~policy over the
pretrained FM expert holds in all four suites and is largest on Long (92.0\% \textit{vs.}\ 81.0\%). Successful episodes of \m{}~policy and of the pretrained
FM expert take a similar number of control steps (156.7 and 156.1).

\begin{table}[t]
\centering
\caption{The $\pi_{0.5}$ VLA per LIBERO suite, success rate (\%) at epoch 1, one seed, with 10 episodes per task. FM starts from the pretrained action expert of $\pi_{0.5}$, and -R marks a randomly initialized one. LIBERO-plus is the shared set of 2008 instances.}
\label{tab:a_vla}
\footnotesize
\begin{tabular}{lcccccc}
\toprule
Method & Spatial & Object & Goal & Long & LIBERO & LIBERO-plus \\
\midrule
\m{} (ours) & \textbf{96.0} & \textbf{100.0} & \textbf{96.0} & \textbf{92.0} & \textbf{96.0} & \textbf{71.5} \\
FM & 90.0 & 91.0 & 90.0 & 81.0 & 88.0 & 69.2 \\
FM-R & 83.0 & 89.0 & 71.0 & 59.0 & 75.5 & 52.6 \\
\bottomrule
\end{tabular}
\end{table}

\subsection{World Action Model}
\label{sec:app_wam}

We fine-tune ImageWAM~\citep{zhang2026imagewam} on the four LIBERO suites jointly (1712 demonstrations) for 10
epochs with one seed. Its image transformer (3.9B parameters) and action transformer are fine-tuned, and the text
encoder is frozen. \m{}~policy keeps the 25 blocks of the action transformer, which read a 512-dimensional
condition once per chunk through a learned query, and replaces its input, time embedding, and output layers with
a FiLM-MLP vector field of four residual layers of width 512. The proprioception enters only the field. The
field integrates chunks of 12 steps with \textit{Euler}, with the contractive perturbation (1\,cm and 0.03\,rad,
applied to half of the training chunks), seamless chunking, and the diversity head, and a separate head predicts the
gripper. The settings that
differ between the heads are listed in Table~\ref{tab:a_wam_setup}. LIBERO-plus uses a shared set of 2008
instances with one episode each and the same seed for all methods, and its success rate is weighted by
perturbation type. All standard deviations in this subsection are over tasks, not over seeds.

\begin{figure}[!tbp]
\centering
\includegraphics{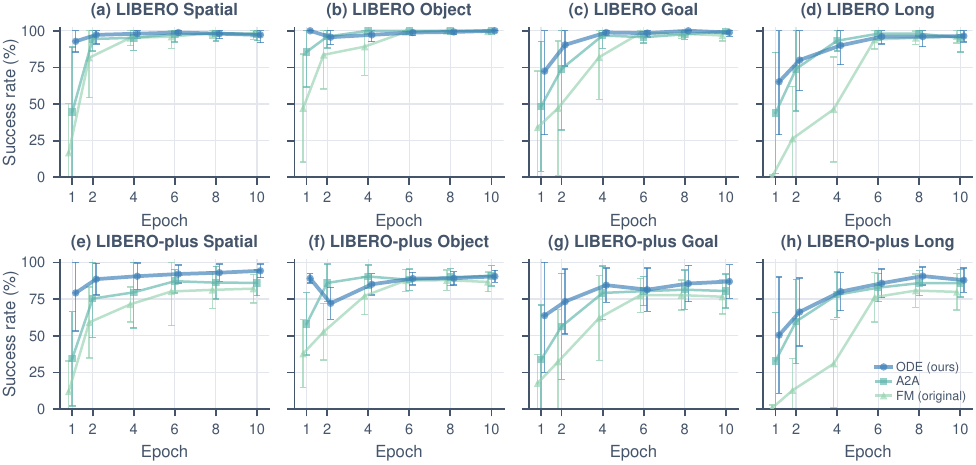}%
\caption{ImageWAM per suite on LIBERO \textbf{(A-D)} and on the shared LIBERO-plus set \textbf{(E-H)}, with
the ODE head (ours), an A2A head, and the original flow matching head (FM). Whiskers show the standard deviation
over the ten tasks of each suite.}
\label{fig:wam_suites}
\end{figure}

Fig.~\ref{fig:wam_suites} shows each suite. The early lead of \m{}~policy is largest on Spatial and Long on LIBERO,
and it lasts until epoch 10 on LIBERO-plus Spatial. Table~\ref{tab:a_wam_plus} lists the success rate per
perturbation type. On the full LIBERO-plus set of 10030 instances, \m{}~policy at epoch 10 succeeds in 88.7\%
of the episodes, close to its 89.9\% on the shared set. The baselines were not evaluated on the full set.

\begin{table}[t]
\centering
\caption{ImageWAM settings that differ between the action heads. All heads use AdamW with a learning rate of $10^{-4}$, a warm-up of half an epoch, and a cosine decay to $10^{-6}$, the current images of two cameras at $224\times224$ without image history, and no EMA, and each epoch is one pass over the same data. Each LIBERO checkpoint is evaluated with 25 episodes per task.}
\label{tab:a_wam_setup}
\footnotesize
\begin{tabular}{p{3.0cm}p{3.0cm}p{3.0cm}p{4.2cm}}
\toprule
\raggedright Setting & \raggedright FM (original) & \raggedright A2A & \raggedright ODE (ours) \tabularnewline
\midrule
\raggedright Action generation & \raggedright 10 flow matching steps & \raggedright 1 flow step from past actions & \raggedright 1 condition read and 12 \textit{Euler} steps \tabularnewline
\raggedright Predicted / executed actions per chunk & \raggedright 16 / 12 & \raggedright 16 / 12 & \raggedright 12 / 12 \tabularnewline
\raggedright Action space & \raggedright controller command & \raggedright controller command & \raggedright desired action, converted to the controller command \tabularnewline
\raggedright Proprioception & \raggedright condition token & \raggedright condition token & \raggedright input of the vector field \tabularnewline
\raggedright Gripper & \raggedright part of the action flow & \raggedright part of the action flow & \raggedright separate head \tabularnewline
\raggedright Trainable parameters & \raggedright 4517.6M & \raggedright 4517.6M & \raggedright 4518.2M \tabularnewline
\bottomrule
\end{tabular}
\end{table}

\begin{table}[t]
\centering
\caption{ImageWAM on LIBERO-plus per perturbation type, success rate (\%) with the standard deviation over tasks. Left, the shared set of 2008 instances at the best checkpoint of each head (epoch 8 for FM and A2A, epoch 10 for ODE). Right, \m{}~policy at epoch 10 on the full set of 10030 instances, per suite.}
\label{tab:a_wam_plus}
\footnotesize
\begin{tabular}{l ccc ccccc}
\toprule
& \multicolumn{3}{c}{Shared set (2008)} & \multicolumn{5}{c}{ODE (ours), full set (10030)} \\
\cmidrule(lr){2-4}\cmidrule(lr){5-9}
Perturbation & FM & A2A & ODE (ours) & Spatial & Object & Goal & Long & All \\
\midrule
Camera & 78.7{\scriptsize$\pm$29.5} & 82.9{\scriptsize$\pm$24.6} & \textbf{91.1}{\scriptsize$\pm$15.2} & 92.6{\scriptsize$\pm$9.5} & 99.2{\scriptsize$\pm$1.0} & 92.2{\scriptsize$\pm$12.2} & 75.7{\scriptsize$\pm$21.2} & 89.7{\scriptsize$\pm$15.2} \\
Robot initial pose & 52.5{\scriptsize$\pm$26.7} & 61.6{\scriptsize$\pm$24.6} & \textbf{71.7}{\scriptsize$\pm$22.8} & 77.1{\scriptsize$\pm$11.8} & 63.3{\scriptsize$\pm$11.4} & 67.0{\scriptsize$\pm$17.0} & 68.7{\scriptsize$\pm$13.3} & 68.8{\scriptsize$\pm$14.2} \\
Language & 89.2{\scriptsize$\pm$17.0} & \textbf{92.1}{\scriptsize$\pm$14.8} & \textbf{92.1}{\scriptsize$\pm$14.2} & 96.7{\scriptsize$\pm$9.8} & 93.8{\scriptsize$\pm$11.1} & 89.0{\scriptsize$\pm$10.6} & 88.8{\scriptsize$\pm$21.1} & 92.0{\scriptsize$\pm$13.6} \\
Light & 97.4{\scriptsize$\pm$5.9} & 95.6{\scriptsize$\pm$16.2} & \textbf{98.7}{\scriptsize$\pm$3.7} & 99.7{\scriptsize$\pm$0.9} & 100.0{\scriptsize$\pm$0.0} & 94.6{\scriptsize$\pm$9.5} & 96.4{\scriptsize$\pm$4.4} & 97.7{\scriptsize$\pm$5.2} \\
Background & 88.5{\scriptsize$\pm$16.6} & \textbf{95.1}{\scriptsize$\pm$12.7} & 93.8{\scriptsize$\pm$16.5} & 95.3{\scriptsize$\pm$15.6} & 96.8{\scriptsize$\pm$4.9} & 94.0{\scriptsize$\pm$12.6} & 91.7{\scriptsize$\pm$11.0} & 94.3{\scriptsize$\pm$11.4} \\
Sensor noise & 92.9{\scriptsize$\pm$11.3} & 95.6{\scriptsize$\pm$10.5} & \textbf{98.3}{\scriptsize$\pm$4.7} & 97.7{\scriptsize$\pm$3.6} & 99.1{\scriptsize$\pm$2.4} & 98.2{\scriptsize$\pm$3.4} & 94.4{\scriptsize$\pm$7.0} & 97.3{\scriptsize$\pm$4.6} \\
Layout & 79.9{\scriptsize$\pm$23.3} & 83.1{\scriptsize$\pm$23.3} & \textbf{86.6}{\scriptsize$\pm$21.8} & 94.3{\scriptsize$\pm$13.1} & 85.9{\scriptsize$\pm$5.4} & 70.4{\scriptsize$\pm$19.6} & 92.9{\scriptsize$\pm$7.2} & 85.1{\scriptsize$\pm$16.1} \\
\midrule
All & 81.9{\scriptsize$\pm$10.8} & 85.8{\scriptsize$\pm$10.3} & \textbf{89.9}{\scriptsize$\pm$8.0} & 93.2{\scriptsize$\pm$4.8} & 90.5{\scriptsize$\pm$2.0} & 85.4{\scriptsize$\pm$7.4} & 86.1{\scriptsize$\pm$7.0} & 88.7{\scriptsize$\pm$6.3} \\
\bottomrule
\end{tabular}
\end{table}

\subsection{Real-World Experiments}
\label{sec:app_real}

We report inference cost and smoothness on the cup task of the dexterous hand in Table~\ref{tab:a_real_cup}. The
baselines never succeed on this task, so their measures come from failed trials and serve only as a reference.
\m{}~policy generates a chunk 3.4 and 5.1 times faster than FM-DiT and DDIM-UNet. Its acceleration at chunk
boundaries is 0.86 times that inside chunks, close to the 0.96 of the demonstrations, whereas the baselines reach 2.03
and 2.97. Its arm jerk is about one third of that of the baselines and below that of the demonstrations, as on the
can task.

\begin{table}[t]
\centering
\caption{Inference cost and smoothness on the real cup task. Inference is measured on one RTX 4090 over all chunks of
all trials. Smoothness is computed over the first 12\,s after the policy takes over, as the mean $\pm$ standard
deviation over trials. The command acceleration is taken over the 14 arm joints, and the jerk of the measured motion over the
14 arm joints and the 22 joints of the right hand. The demonstrations are the 50 training demonstrations from
SharpaDex, cut into chunks of 8 steps.}
\label{tab:a_real_cup}
\footnotesize
\begin{tabular}{lcccc}
\toprule
 & Demonstrations & \m{}~policy (ours) & FM-DiT & DDIM-UNet \\
\midrule
Time per chunk (ms) & -- & \textbf{10.2} & 35.0 & 51.8 \\
95th percentile and maximum time (ms) & -- & \textbf{12.1} and 65.8 & 36.6 and 49.5 & 53.6 and 65.2 \\
Acceleration at chunk boundaries (rad/s$^2$) & 1.04 & $\mathbf{0.29\pm0.06}$ & $6.40\pm0.82$ & $10.9\pm4.5$ \\
Acceleration inside chunks (rad/s$^2$) & 1.09 & $\mathbf{0.34\pm0.06}$ & $3.15\pm0.10$ & $4.02\pm2.30$ \\
Boundary to inside ratio & 0.96 & \textbf{0.86} & 2.03 & 2.97 \\
Arm jerk (rad/s$^3$) & 10.3 & $\mathbf{6.1\pm0.6}$ & $19.6\pm1.7$ & $18.1\pm1.4$ \\
Hand jerk (rad/s$^3$) & 47 & $\mathbf{37\pm15}$ & $62\pm8$ & $49\pm6$ \\
\bottomrule
\end{tabular}
\end{table}

The pill task in Table~\ref{tab:a_real_pill} gives the same picture with a baseline that also succeeds.
\m{}~policy generates a chunk 3.5 times faster than FM-DiT. Its ratio of the acceleration at chunk boundaries to that
inside chunks is 1.45, close to the 1.24 of the demonstrations, against 2.17 for FM-DiT. Its arm jerk is half of that
of FM-DiT and below that of the demonstrations, while the hand jerk of both methods is close and below that of the
demonstrations.

\begin{table}[t]
\centering
\caption{Inference cost and smoothness on the real pill task. Inference is measured on one RTX 4090 over all chunks of
all trials. Smoothness is computed over the first 12\,s after the policy takes over, as the mean $\pm$ standard
deviation over the 17 successful trials of \m{}~policy and all 20 trials of FM-DiT. The command acceleration is taken
over the 14 arm joints, and the jerk of the measured motion over the 14 arm joints and the 22 joints of the right
hand. The demonstrations are the 50 training demonstrations from SharpaDex, cut into chunks of 8 steps.}
\label{tab:a_real_pill}
\footnotesize
\begin{tabular}{lccc}
\toprule
 & Demonstrations & \m{}~policy (ours) & FM-DiT \\
\midrule
Time per chunk (ms) & -- & \textbf{10.3} & 36.3 \\
95th percentile and maximum time (ms) & -- & \textbf{12.7 and 30.5} & 42.6 and 89.4 \\
Acceleration at chunk boundaries (rad/s$^2$) & 1.57 & $\mathbf{1.14\pm0.15}$ & $6.12\pm0.58$ \\
Acceleration inside chunks (rad/s$^2$) & 1.27 & $\mathbf{0.78\pm0.07}$ & $2.83\pm0.20$ \\
Boundary to inside ratio & 1.24 & \textbf{1.45} & 2.17 \\
Arm jerk (rad/s$^3$) & 12.8 & $\mathbf{9.8\pm0.6}$ & $19.6\pm1.5$ \\
Hand jerk (rad/s$^3$) & 47 & $\mathbf{30\pm3}$ & $31\pm4$ \\
\bottomrule
\end{tabular}
\end{table}

\subsection{Reinforcement Learning}
\label{sec:app_rl}

We build on an open-source reproduction of SuperMimic~\citep{liao2026supermimic}. Each iteration steps 4096
simulated environments once and stores the transitions, trains the world model on them, and updates the actor
by backpropagating the discounted return over 32 steps imagined in the frozen world model, plus the value of the
critic at the last step. The task is whole-body tracking of the 40 complete LAFAN1 clips (2.45 hours at 50\,Hz,
100 to 300\,s each) with a Unitree G1 humanoid with 29 actuated joints. The action is the offset of the
joint-position targets from the default pose, tracked by a PD controller. The actor encodes the target motion into
a latent $\mathbf{z}_t\in\mathbb{R}^{32}$ and decodes it with 10 frames of proprioception into the action, with
8.91M parameters. We add the anchor-relative pose of BeyondMimic~\citep{liao2026beyondmimic} to the goal for both
actors, since with proprioception alone the heading drifts over such long clips and the return saturates at about
0.58.

The ODE actor keeps this network and its initialization and only reads the decoder output as the velocity in
Eq.~\eqref{eq:rl_step}, where the previous action is the noise-free output of the actor, so exploration noise does
not accumulate along the integration. Without bounds, the integration turned an early error of the world model
into a drift of the action and a collapse of the return in one of the runs. We therefore bound the change per step
by 1.62 and each joint of the action by its own limit. Both bounds are set slightly above the 99th and 99.9th
percentiles of the actions of converged policies of both actors, so they are rarely active.

Everything else, including the world model, the critic, the rewards, the training noise, and the hyperparameters,
follows the reproduction. Each run trains for 3 hours on one NVIDIA H200 GPU, about 33k iterations. Following
SuperMimic, every clip is played once from its first frame with the deterministic policy and without noise or
randomization, and an episode ends when the pelvis deviates from the reference beyond a threshold. We report the
reward per step normalized by its maximum, averaged over all steps of each clip with the steps after a fall counted
as zero and then over the 40 clips, and the mean squared second difference of the executed actions.

\begin{figure}[!tbp]
\centering
\includegraphics{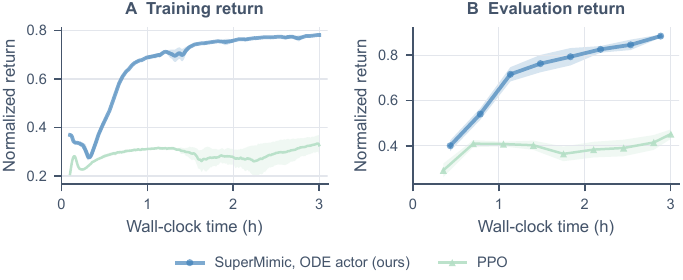}%
\caption{\textbf{Returns in reinforcement learning.} The ODE actor of SuperMimic \textit{vs.}\ PPO on whole-body
tracking of the 40 LAFAN1 clips, over wall-clock time on one GPU. \textbf{(A)} Training return.
\textbf{(B)} Normalized return on all 40 clips with the deterministic policy, where the steps after a fall count as
zero. Lines and shaded bands show the mean and standard deviation over three random seeds.}
\label{fig:a_rl_return}
\end{figure}

PPO~\citep{schulman2017ppo} uses the same environment, observations, rewards, termination conditions, domain
randomization, and clips as SuperMimic, but no world model. It samples its own exploration actions, so the
execution noise of the environment is turned off. Its hyperparameters follow the G1 tracking configuration of
BeyondMimic~\citep{liao2026beyondmimic}, with 8192 parallel environments and the network size of SuperMimic. Each
run also trains for 3 hours on one H200 GPU. Since an iteration of PPO differs from one of SuperMimic, all curves are
plotted over wall-clock time. Fig.~\ref{fig:a_rl_return} compares the returns of the ODE actor and PPO. The training return
of PPO includes its own exploration noise and is shown only for reference.

\subsection{Ablation Studies}
\label{sec:app_ablation}

\textbf{{Diversity head.}} The head reads the features of the field with the gradient stopped, and its loss
does not reach the field, so the field receives exactly the same gradient with or without the head
(Section~\ref{sec:method_side}). The head thus leaves learning unchanged and only alters the actions at
inference. We evaluate the same models of three seeds without the offset ($\boldsymbol{\epsilon}=\mathbf{0}$, i.e., the
plain ODE), with the mean of the drawn component, and with full sampling within the component
(Table~\ref{tab:e12_div}). After one epoch, when the head is not yet trained, the offset lowers the success
rate from $85.3\pm1.9$\% to $79.8\pm1.6$\% with component means and to $52.1\pm0.1$\% with full sampling. From epoch 5 on, the three
differ by at most 4.2 points, and full sampling even reaches the highest $94.8\pm0.4$\% at epoch 20. The diversity
head therefore mainly affects the early performance, which component means largely preserve, and we use
component means for uni-modal tasks.

\begin{table}[t]
\centering
\caption{Ablation on the vector field architecture, with the DiT baselines as references. Both ODE variants use RK4 with $H=8$. Success rate (\%) on LIBERO, averaged over the four suites (mean over three random seeds, except ODE-DiT with one seed, each suite with 10 tasks $\times$ 20 episodes), training speed (iterations per second, batch size 32), and inference time per chunk of 8 actions (ms, shared evaluation GPUs). The two \m{} variants have matched parameters outside the encoder.}
\label{tab:arch}
\footnotesize
\begin{tabular}{llccccc}
\toprule
Method & Vector field & Ep.~1 & Ep.~10 & Ep.~100 & it/s & ms/chunk \\
\midrule
ODE-MLP (ours) & FiLM-MLP & \textbf{79.2} & \textbf{93.9} & \textbf{93.8} & 17.1 & \underline{11.0} \\
ODE-DiT (ours) & AdaLN Transformer & \underline{77.2} & \underline{89.6} & \underline{91.2} & 3.5 & 51.9 \\
\midrule
FM-DiT & -- & 13.5 & 80.1 & 85.4 & \textbf{31.6} & 61.4 \\
A2A-DiT & -- & 42.2 & 86.9 & 85.6 & \underline{23.3} & \textbf{5.8} \\
\bottomrule
\end{tabular}
\end{table}

\textbf{{Vector field architecture.}} We replace the FiLM-MLP vector field with AdaLN Transformer blocks of
FM-DiT, keeping the number of parameters (ODE-DiT, Table~\ref{tab:arch}). With the Transformer,
\m{}~policy still outperforms all baselines, with 77.2\% after one epoch and 91.2\% at epoch 100 (one seed),
against at most 43.2\% and 87.8\%, and it exceeds FM-DiT with the same blocks by 5.8 points at epoch 100. However,
it stays below the MLP field at every checkpoint and is much slower, training at 3.5 instead of 17.1
iterations per second and taking 51.9 instead of 11.0\,ms per chunk (Table~\ref{tab:a_arch_exec}). Two reasons explain this. First, RK4
evaluates the field 32 times per chunk, in training as well as in execution, so the overhead of each
Transformer call is multiplied. Second, the Transformer of FM-DiT models the dependencies among the actions
of a chunk, whereas in \m{}~policy the integration already produces the chunk step by step, and the field
only maps the current state to its velocity, a low-dimensional mapping that an MLP fits well. A simple MLP
field is therefore sufficient, as the integration carries the temporal structure of the chunk.

\begin{table}[t]
\centering
\caption{The diversity head at inference, for the same models. Without the offset ($\boldsymbol{\epsilon}=\mathbf{0}$, the plain ODE), with the mean of the drawn component, and with full sampling within the component. Success rate (\%) averaged over the four LIBERO suites, mean$\pm$std over three random seeds, with the per-suite means in parentheses (Object / Spatial / Goal / Long), and 10 episodes per task.}
\label{tab:e12_div}
\footnotesize
\setlength{\tabcolsep}{3pt}
\begin{tabular}{lcccc}
\toprule
Inference & Epoch 1 & Epoch 5 & Epoch 10 & Epoch 20 \\
\midrule
No offset & \textbf{85.3}{\scriptsize$\pm$1.9} {\scriptsize(98/81/91/72)} & \textbf{95.0}{\scriptsize$\pm$1.0} {\scriptsize(99/97/96/87)} & 92.8{\scriptsize$\pm$1.0} {\scriptsize(99/94/93/85)} & 93.1{\scriptsize$\pm$1.2} {\scriptsize(99/97/93/84)} \\
Component mean (ours) & 79.8{\scriptsize$\pm$1.6} {\scriptsize(98/78/83/60)} & 94.8{\scriptsize$\pm$0.4} {\scriptsize(100/97/96/87)} & \textbf{94.3}{\scriptsize$\pm$1.3} {\scriptsize(99/98/95/86)} & 94.2{\scriptsize$\pm$1.0} {\scriptsize(99/95/95/87)} \\
Full sampling & 52.1{\scriptsize$\pm$0.1} {\scriptsize(79/50/52/28)} & 90.8{\scriptsize$\pm$0.0} {\scriptsize(96/94/90/82)} & 93.7{\scriptsize$\pm$1.1} {\scriptsize(98/93/96/87)} & \textbf{94.8}{\scriptsize$\pm$0.4} {\scriptsize(99/99/94/87)} \\
\bottomrule
\end{tabular}
\end{table}

\begin{table}[t]
\centering
\caption{Ablation on the vector field architecture. Both ODE variants use RK4 with $H=8$. Execution quality and inference time on LIBERO, over the successful episodes of all suites and seeds (mean $\pm$ standard deviation over episodes). SR is the mean and standard deviation of the success rate over three random seeds. In-chunk and cross-chunk denote the mean change of the command between consecutive steps inside a chunk and at the first step of a new chunk. Jerk is the RMS jerk of the measured end-effector position (m/s$^3$). Steps is the number of control steps of a successful episode, and time is the inference time per action and per call (ms) of one seed, measured on shared evaluation GPUs. ODE-DiT is trained with one seed.}
\label{tab:a_arch_exec}
\footnotesize
\setlength{\tabcolsep}{3.5pt}
\begin{tabular}{llccccccc}
\toprule
Method & Ep. & SR & In-chunk & Cross-chunk & Jerk & Steps & ms/action & ms/call \\
\midrule
ODE-MLP (ours) & 20 & \textbf{92.9}{\scriptsize$\pm$0.5} & $\mathbf{0.01871}${\scriptsize$\pm$0.0051} & $\mathbf{0.02855}${\scriptsize$\pm$0.0059} & $\mathbf{5.257}${\scriptsize$\pm$2.1} & $\mathbf{146.5}${\scriptsize$\pm$71} & $\mathbf{1.43}${\scriptsize$\pm$0.079} & $\mathbf{11.11}${\scriptsize$\pm$0.55} \\
 & 50 & \textbf{93.8}{\scriptsize$\pm$0.3} & $\mathbf{0.0196}${\scriptsize$\pm$0.0053} & $\mathbf{0.0257}${\scriptsize$\pm$0.0054} & $\mathbf{5.577}${\scriptsize$\pm$2.3} & $\mathbf{146}${\scriptsize$\pm$70} & $\mathbf{1.432}${\scriptsize$\pm$0.072} & $\mathbf{11.15}${\scriptsize$\pm$0.56} \\
 & 100 & \textbf{93.8}{\scriptsize$\pm$1.3} & $\mathbf{0.02017}${\scriptsize$\pm$0.0054} & $\mathbf{0.02453}${\scriptsize$\pm$0.0053} & $\mathbf{5.736}${\scriptsize$\pm$2.3} & $147.2${\scriptsize$\pm$72} & $\mathbf{1.414}${\scriptsize$\pm$0.069} & $\mathbf{11}${\scriptsize$\pm$0.47} \\
\midrule
ODE-DiT (ours) & 20 & 89.4 & $0.01914${\scriptsize$\pm$0.005} & $0.02893${\scriptsize$\pm$0.0057} & $5.276${\scriptsize$\pm$2.1} & $148.5${\scriptsize$\pm$73} & $7.192${\scriptsize$\pm$0.23} & $55.97${\scriptsize$\pm$1.3} \\
 & 50 & 90.8 & $0.02058${\scriptsize$\pm$0.0052} & $0.02596${\scriptsize$\pm$0.0059} & $5.734${\scriptsize$\pm$2.3} & $146.6${\scriptsize$\pm$73} & $7.155${\scriptsize$\pm$0.26} & $55.63${\scriptsize$\pm$1.5} \\
 & 100 & 91.2 & $0.02133${\scriptsize$\pm$0.0053} & $0.02516${\scriptsize$\pm$0.0057} & $5.942${\scriptsize$\pm$2.3} & $\mathbf{144.5}${\scriptsize$\pm$69} & $6.672${\scriptsize$\pm$0.19} & $51.91${\scriptsize$\pm$0.92} \\
\bottomrule
\end{tabular}
\end{table}

\end{document}